\documentclass[preprint,12pt]{elsarticle}
\usepackage{textcomp}

\usepackage{amssymb}
\usepackage{amsmath}
\usepackage{titlesec}
\titleformat{\paragraph}[runin]
{\normalfont\bfseries}
{\theparagraph}{1em}{}

\usepackage{hyperref}
\usepackage{algorithm}
\usepackage{algpseudocode}
\usepackage{booktabs}      
\usepackage{multirow}      
\usepackage{adjustbox}     
\usepackage{float}
\usepackage{amsmath}
\usepackage{subcaption}
\usepackage{xcolor}
\usepackage{multirow}

\journal{Journal of Biomedical Informatics}
\titleformat{\section}
  {\large\bfseries}{\thesection}{2em}{}

\titleformat{\subsection}
  {\normalsize\bfseries}{\thesubsection}{1.5em}{}

\titleformat{\subsubsection}
  {\small\bfseries}{\thesubsubsection}{0.5em}{}

\titlespacing*{\section}{0pt}{6pt}{6pt}
\titlespacing*{\subsection}{0pt}{4pt}{4pt}
\titlespacing*{\subsubsection}{0pt}{2pt}{2pt}

\titleformat{\paragraph}[runin]{\normalfont\bfseries}{\theparagraph}{1em}{} 
\begin{document}

\begin{frontmatter}



\title{ARMA-C$^3$: A \underline{C}ontrastive ARMA \underline{C}onvolutional Framework for Unsupervised and Semi-supervised \underline{C}lassification}

%

\author[1]{VSS Tejaswi Abburi}
\author[1]{Saurabh J. Shigwan}
\author[1]{Nitin Kumar}

\affiliation[1]{organization={Department of Computer Science and Engineering, Shiv Nadar Institute of Eminence Deemed to be University},
               city={Delhi NCR},
               country={India}}

\begin{abstract}
In biomedical and neurodegenerative disorders, accurate and early disease identification is still difficult, partly because of the scarcity of labeled data and the complexity of imaging patterns. To address these challenges, we introduce \textbf{ARMA-C$^3$}, a unified unsupervised and semi-supervised graph learning framework for node classification based on contrastive learning and graph-cut regularization to learn structurally meaningful and discriminative representations. By modeling samples or images as graph nodes and exploiting inter-sample relationships, the proposed framework leverages subject-level relationships that conventional machine learning methods typically overlook. We perform extensive binary classification experiments across five clinically relevant datasets: Alzheimer’s Disease Neuroimaging Initiative (ADNI), the Neuroimaging in Frontotemporal Dementia (NIFD) dataset,three medical imaging benchmarks (BreastMNIST, PneumoniaMNIST, and a liver ultrasound dataset)  Experimental results demonstrate that ARMA-C$^3$ achieves competitive and frequently superior performance compared to classical clustering techniques, state-of-the-art machine learning models, and existing graph-based deep learning approaches across multiple evaluation settings, particularly under limited supervision and severe class imbalance. The proposed framework further demonstrates robust representation learning and strong cross-modal generalization across diverse biomedical imaging modalities.\\
    *The Github repository of the code is available on \href{https://github.com/tejaswi-abburi1083/ARMAC3-2026.git}{ARMA-C$^3$}
\end{abstract}



\end{frontmatter}
\section{Introduction}
\label{sec:intro}

Neurodegenerative and medical disorders represent a major challenge for global healthcare systems. This has created a growing need for reliable, data-driven methods for early diagnosis, disease characterization, and patient management. Among these conditions, Alzheimer's disease (AD) and its early stage, Mild Cognitive Impairment (MCI), are particularly important because of their progressive nature and the potential benefits of early treatment. Likewise, Frontotemporal Dementia (FTD) is a major cause of early-onset dementia, featuring varied clinical symptoms that complicate diagnosis. In addition to neurological disorders, tasks such as breast lesion assessment, liver characterization, and pneumonia detection also require robust computational methods that work well across different biomedical domains.

Recent progress in neuroimaging and medical imaging, together with advances in artificial intelligence, has opened new opportunities for improved disease analysis across diverse clinical modalities.  In addition to neuroimaging data, large-scale benchmark datasets such as BreastMNIST and PneumoniaMNIST \cite{yang2021medmnist} have demonstrated the growing role of data-driven approaches in breast cancer screening and pulmonary disease diagnosis. In particular, diffusion Magnetic Resonance Imaging (dMRI) can sensitively characterize white-matter microstructural integrity and brain connectivity changes linked to neurodegeneration. It can capture subtle disconnection patterns that show up before obvious atrophy occurs~\cite{caso2016insights,acosta2014diffusion,kantarci2011diffusion}. However, high-dimensional imaging modalities—including dMRI, structural MRI, mammography, and X-ray—are often affected by noise and limited sample sizes. Traditional machine learning models usually treat subjects or images as separate instances, ignoring crucial relational information and subject-level structure that may indicate disease progression.

Graph Neural Networks (GNNs) \cite{scarselli2008graph} present a valuable approach to overcome these issues by modeling subjects as nodes in a graph. They take advantage of relationships between subjects through neighborhood aggregation and shared representation. This relational learning strategy is particularly useful in biomedical applications, where disease symptoms often develop gradually in cohorts instead of appearing as isolated cases~\cite{jack2013tracking}. Subtle shared imaging and structural patterns can indicate clinically meaningful subgroups~\cite{varol2017hydra}.

Despite this promise, most current GNN-based biomedical methods mainly focus on supervised settings and disease-specific applications. These methods often require large amounts of labeled data, which can be expensive to obtain. Moreover, fully exploiting the underlying data structure in the unsupervised and semi-supervised setup remains a challenge.
To address these challenges, we propose ARMA-C$^3$, a unified graph 
learning framework based on spectral filtering with contrastive 
consistency for robust biomedical representation learning. By modeling 
biomedical data as graph signals over subject-level population graphs 
\cite{shuman2013emerging, ortega2018graph}, the proposed approach 
leverages ARMA-based spectral filtering \cite{bianchi2021graph} to 
capture multi-scale relational patterns across heterogeneous imaging 
modalities. While MERIT \cite{jin2021multi} was originally developed 
for general-purpose graph representation learning on citation and social 
networks, it was neither designed nor evaluated for subject-level 
biomedical population graphs — where nodes represent clinical subjects, 
graph structure encodes inter-subject imaging similarity, and label 
scarcity is a fundamental operational constraint. Similarly, prior 
approaches apply structural regularization \cite{chen2020iterative} and 
contrastive learning \cite{jin2021multi} as independent training 
objectives, leaving a critical gap: neither component alone can 
simultaneously preserve community-level cohort structure and enforce 
augmentation-invariant subject representations under noisy graph 
construction. ARMA-C$^3$ closes this gap by jointly optimizing 
modularity-based structural regularization and multi-view contrastive 
consistency within a single unified objective — a coupling that, to the 
best of our knowledge, is novel in the context of biomedical population 
graph learning. This joint formulation enables the learned 
representations to simultaneously preserve graph topology, enhance 
inter-subject discrimination, and improve robustness to noisy graph 
construction.


\textbf{The contributions are as follows: }
\begin{enumerate}
\item We perform fully unsupervised node-level clustering using contrastive graph learning with graph-cut-based structural regularization to learn structurally meaningful and discriminative representations. This uncovers stable subject groups without relying on diagnostic labels, enabling robust disease classification in label-scarce and imbalanced-labeled real-world settings, which previous methods do not fully address.
  \item We demonstrate that the proposed framework generalizes across fundamentally different biomedical modalities—including diffusion MRI (neuroimaging), breast ultrasound, liver ultrasound, and chest X-ray,
  showing strong robustness, adaptability, and cross-modal generalization capability across heterogeneous biomedical modalities.
    \item By leveraging clinically informed and neuroanatomically relevant ROIs, we construct a compact diffusion MRI representation that preserves disease-sensitive microstructural information, thereby improving discriminability between healthy and diseased cohorts.
    

\end{enumerate}

\section{Related Works}
There has been recent exploration of a wide range of machine learning and graph learning methods for solving both clustering and disease prediction challenges in neuroimaging and broader biomedical datasets ~\cite{wen2020convolutional} ~\cite{parisot2018disease}. Traditional unsupervised techniques such as K-means demonstrate limited capability in capturing complex disease patterns, as they rely solely on feature-space distances without modeling graph structure or inter-subject relationships, thereby failing to exploit underlying population organization~\cite{macqueen1967kmeans}. Spectral clustering-based approaches attempt to address this limitation by constructing affinity graphs and leveraging Laplacian properties to derive more meaningful partitions. SpectralNet integrates spectral embedding into a neural framework~\cite{shaham2018spectralnet}, TokenCut computes the Fiedler vector of the normalized Laplacian to perform principled graph cuts~\cite{wang2023tokencut}, and TANGO introduces typicality-aware nonlocal mode seeking followed by graph-cut optimization to enhance robustness and separability~\cite{ma2024tango}. However, these approaches largely depend on fixed affinity graphs with non-learnable embeddings, which may limit adaptability to heterogeneous biomedical populations.
More recently, MAGI~\cite{liu2024revisiting} reformulates modularity maximization through a contrastive learning framework, where community structure naturally guides the formation of positive and negative pairs.
Notably, MAGI collapses modularity and contrastive learning into a single objective by treating community structure itself as the source of positive and negative pairs. In contrast, ARMA-C\textsuperscript{3} retains modularity-based structural regularization and multi-view contrastive consistency as two separate, jointly optimized loss terms; this decoupling allows the contrastive objective to enforce augmentation-invariant representations independently of the cluster geometry imposed by the modularity term, which we hypothesize is the source of ARMA-C\textsuperscript{3}'s improved robustness on imbalanced and noisy biomedical graphs.

Graph Neural Network (GNN)-based methods refine node embeddings through message passing, enabling the modeling of higher-order dependencies and subtle disease-driven structure~\cite{scarselli2008graph}. Graph Convolutional Networks (GCN)~\cite{kipf2017semi} and Graph Attention Networks (GAT)~\cite{velickovic2018graph} learn neighborhood-aware features through layered aggregation, whereas Auto-Regressive Moving Average (ARMA)-based graph convolutional networks~\cite{bianchi2021graph} employ recursive filter stacks to approximate spectral graph convolutions, enabling flexible frequency responses, improved oversquashing control through iterative message refinement, and enhanced multi-hop information propagation by capturing both local and long-range graph dependencies. These developments motivate the use of GNNs in neuroimaging, where relational organization and subtle patterns of disease progression are critical.

Recent advances in biomedical graph learning have further explored graph contrastive learning, population graph construction, and self-supervised representation learning for neuroimaging and medical image analysis. Luo et al.~\cite{luo2024interpretable} propose an interpretable brain-graph contrastive learning framework for brain disorder analysis, but operate on \emph{intra-subject} ROI connectivity graphs rather than inter-subject population graphs, and therefore do not exploit cohort-level relational structure. Chen et al.~\cite{chen2024self} introduce a self-supervised approach with adaptive graph structure and function representation for cross-dataset brain disorder diagnosis, yet rely on modality-specific functional connectivity features that limit transferability beyond neuroimaging. Meng et al.~\cite{meng2024interpretable} construct an interpretable population graph for identifying rapid Alzheimer's progression in the UK Biobank cohort, but operate in a fully supervised regime that presupposes the availability of large-scale longitudinal labels. While these works collectively demonstrate the growing importance of relational subject-level modeling, they remain dependent on predefined graph structures, modality-specific assumptions, or supervised objectives — limiting robustness and generalization across heterogeneous biomedical imaging settings. In contrast, ARMA-C\textsuperscript{3} jointly optimizes modularity-based structural regularization and multi-view contrastive consistency within a single unsupervised objective, and demonstrates consistent generalization across diffusion MRI, ultrasound, and chest X-ray modalities.


Graph-based semi-supervised learning methods, including APPNP
~\cite{gasteiger2018predict} and JacobiConv~\cite{wang2022powerful}, attempt to leverage relational information through neighborhood propagation; however, their dependence on fixed similarity graphs implies inaccuracies in graph construction which may lead to oversmoothing, loss of discriminability, and unreliable subject separation. Although these approaches often outperform purely feature-based models, capturing the heterogeneous and subtle patterns of neurodegenerative progression remains challenging.

In parallel, contrastive graph learning strategies such as Deep Graph Infomax~\cite{velickovic2019deep} have demonstrated promise by learning structure-aware embeddings and improving discrimination capability in both unsupervised and semi-supervised contexts. Building upon this direction, MERIT contrastive learning~\cite{jin2021multi} further enhances representational stability by enforcing augmentation-invariant consistency through teacher–student alignment ~\cite{hinton2015distilling}, yielding more robust embeddings in heterogeneous neuroimaging populations. 


\section{Proposed Method}
\label{Method}
We propose ARMA-C$^3$, a unified graph signal learning framework designed to jointly address structural consistency, representation robustness, and noise sensitivity in biomedical graphs. We call the framework ARMA-C$^3$, where C$^3$ denotes the joint use of Contrastive learning, ARMA Convolution, and graph-Cut regularization. Unlike conventional approaches that apply graph neural networks, structural regularization ~\cite{chen2020iterative} and contrastive learning~\cite{jin2021multi} independently, our framework deploys a contrastive learning framework leveraging graph cut \textbf{within a single optimization formulation.}
In the proposed method, each data point is transformed into compact subject-level features. These features are then used to construct a graph capturing inter-subject similarities. A graph neural network learns node embeddings through message passing, where each node iteratively aggregates information from its neighbors to update its representation. Modularity~\cite{tsitsulin2023graph} constraints encourage a well-separated cohort structure
(section~\ref{sec:struct_reg}). Finally, incorporating MERIT-based contrastive learning~\cite{jin2021multi} ensures multi-view consistency, leading to stable and augmentation-invariant representations for further analysis (section~\ref{sec:cont_learn}).
\subsection{Preprocessing and Feature Extraction}
This stage aims to change raw biomedical data into standardized subject representations that work well for graph learning. Depending on the dataset type, we extract either diffusion MRI-based neuroimaging descriptors or pretrained feature embeddings. We normalize these specific features and combine them to create subject-wise vectors. These vectors will later be used as node attributes in the graph we build.
For the diffusion MRI cohorts, all scans $\{d_i\}_{i=1}^{n}$ were first converted from DICOM to NIFTI format, followed by diffusion tensor estimation and computation of Fractional Anisotropy (FA) maps. Each FA map was nonlinearly registered to the JHU white-matter atlas to obtain subject-specific ROI parcellations. Mean FA values are extracted from $p$ predefined ROIs $\{r_j\}_{j=1}^{p}$ (details in Section~\ref{sec:exp_res_disc}). 

Subsequently, we construct histogram-based representations of FA distributions within each ROI.  For each subject $s_i$ and each ROI $r_j$, we collect the FA values 
$\mathcal{V}_{i,j} \in [0,1]^{k_{ij}}$,
where $k_{ij}$ denotes the number of voxels within that region. The interval $[0,1]$ is partitioned into $q$ equal-width bins 
$\mathcal{B}=\{b_1,b_2,\ldots,b_q\}$,
and a normalized histogram is computed as
\[
h_{i,j} = \frac{\mathrm{Histogram}(\mathcal{V}_{i,j}, \mathcal{B})}{k_{ij}},
\]
yielding an empirical FA distribution descriptor for each ROI. Concatenating all ROI-wise histograms forms a subject-level feature vector $H_{s_i}$. Collectively, we obtain
\[
X = [H_{s_i}]_{i=1}^{n} \in \mathbb{R}^{n \times pq},
\]
which serves as the node feature matrix in the constructed graph. 
For image-based datasets, each sample was normalized to $[0,1]$ and encoded
using a pretrained deep visual backbone to obtain compact feature
representations. Specifically, BreastMNIST ultrasound images $I_i$ and liver Ultrasound images $L_i$ are processed using a pretrained Vision Transformer (ViT-DINO)~\cite{caron2021emerging}, while PneumoniaMNIST chest X-ray images $J_i$ are channel-replicated and encoded using a pretrained ResNet-18 model (truncated before the classifier)~\cite{he2016deep}. Since end-to-end optimization of large pretrained encoders such as ViT-DINO and ResNet-18 incurs substantial computational overhead and increases the risk of overfitting in limited-data biomedical settings, we adopt frozen feature extraction as a practical, computationally efficient design strategy. All input images are resized to \(224 \times 224\) to maintain compatibility with the pretrained encoder architectures and weight configurations. The resulting image embeddings were stacked to
form the node feature matrix $X \in \mathbb{R}^{n \times d}$, which serves as
input to the graph learning framework.

\medskip
\noindent
This strategy yields compact pretrained feature embeddings for all datasets,
which serve as node attributes for subsequent graph learning.
After extracting features, each subject is represented by a $d$-dimensional vector 
$\mathbf{x}_i \in \mathbb{R}^{d}$. We then create a subject-level graph
\[
G = (V,E,X),
\]
where each node represents a subject, and 
$X = [\mathbf{x}_1,\ldots,\mathbf{x}_n]^{\top} \in \mathbb{R}^{n\times d}$.

\noindent\textbf{\footnotesize Adjacency Construction:}
We encode inter-subject similarity using cosine similarity measure. Each feature vector is 
$\ell_2$-normalized. We compute pairwise similarity as follows:
\[
W_{ij}=\frac{\mathbf{x}_i^{\top}\mathbf{x}_j}{\|\mathbf{x}_i\|_2\|\mathbf{x}_j\|_2},
\]
and then apply a threshold $\alpha$ as follows:
\[
A_{ij}=
\begin{cases}
W_{ij}, & W_{ij}\ge\alpha,\\[2pt]
0, & \text{otherwise}.
\end{cases}
\]
Here, $\alpha$ acts as a sparsity controller: higher values retain only strong, reliable similarities and suppress noisy edges, while lower values preserve broader connectivity. Thus, $\alpha$ balances noise reduction with sufficient graph structure for effective message passing. This results in a sparse, undirected weighted adjacency matrix $A\in[0,1]^{n\times n}$.

This sparsification strategy acts as an implicit denoising mechanism, promoting local smoothness of graph signals and improving the robustness of downstream learning. Moreover, although the graph structure is fixed, the joint use of contrastive learning and structural regularization progressively refines node representations, mitigating the impact of initial graph noise.

\noindent\textbf{\footnotesize Graph Representation for Learning:}
For efficient training, $A$ is converted into a sparse edge-index form that keeps only
non-zero edges as $(i,j)$ pairs, reducing both memory usage and computation:
\[
E = \{(i,j)\,:\,A_{ij}>0\}.
\]
This results in the final graph $G=(V,E,X,A)$. Here, $A$ supports structural
regularization (section~\ref{sec:struct_reg}) while $(E,X)$ enable
message passing in the GNN. This graph serves as the foundation for the
next learning framework.
\subsection{Graph Neural Network Learning Framework}

Our goal is to learn distinct node
embeddings that reflect both local neighborhood structure and the overall
graph organization. Unlike conventional unsupervised GNNs that rely solely on message passing, our
framework combines structural regularization with contrastive
learning. Structural regularization, via modularity maximization ~\cite{tsitsulin2023graph}, imposes explicit constraints on the learned cluster assignments, preserving meaningful
cluster structures and preventing feature mixing, where repeated message passing causes embeddings from different clusters to mix, making them harder to distinguish and leading to oversmoothing.
Complementarily, MERIT-based contrastive learning~\cite{jin2021multi} encourages representation consistency by aligning node embeddings obtained from multiple augmented graph views through a teacher–student framework~\cite{hinton2015distilling}, thereby reducing sensitivity to local feature noise and global structural perturbations.
Together, these components yield stable and well-separated embeddings that
generalize effectively across biomedical datasets.
The detailed design of the encoder, objectives, and contrastive formulation are outlined in the following subsections.
\subsubsection{Contrastive Learning}
\label{sec:cont_learn}

To enhance the robustness and discriminability of learned graph representations
in \textbf{ARMA-C$^3$}, we adopt the Multi-scale Contrastive Siamese Network
(MERIT)~\cite{jin2021multi}, a state-of-the-art contrastive learning framework
tailored for graph-structured data. By enforcing consistency across perturbed
graph views, MERIT mitigates the influence of noisy edges that can otherwise
misguide information propagation in message-passing GNNs. Through a
teacher-student paradigm, MERIT aligns representations across multiple
augmented views of the same graph, producing stable target embeddings. This
improves robustness to local neighborhood perturbations (e.g., feature masking
and edge dropping) as well as global structural variations.

The objective combines cross-view consistency and cross-network teacher--student alignment losses to enforce representation consistency between two augmented graph views:
\begin{align*}
\mathcal{L}_1 &= \beta \mathcal{L}_{\mathrm{wo\text{-}cross\text{-}view}}(\mathbf{h}_1,\mathbf{h}_2) +(1-\beta)\mathcal{L}_{\mathrm{wo\text{-}cross\text{-}network}}(\mathbf{h}_1,\mathbf{z}_2) \\
\mathcal{L}_2 &= \beta \mathcal{L}_{\mathrm{wo\text{-}cross\text{-}view}}(\mathbf{h}_2,\mathbf{h}_1) + (1-\beta)\mathcal{L}_{\mathrm{wo\text{-}cross\text{-}network}}(\mathbf{h}_2,\mathbf{z}_1) \\
\mathcal{L}_{\mathrm{con}} &= \frac{\mathcal{L}_1 + \mathcal{L}_2}{2}
\end{align*}
Here, $\mathbf{h}_1,\mathbf{h}_2$ are the outputs of an ARMA-based graph neural
network encoder and $\mathbf{z}_1,\mathbf{z}_2$ are target representations produced by the Siamese network. $\beta \in [0,1]$ controls the relative contribution of 
(i) \emph{intra-network consistency} $\mathcal{L}_{\mathrm{wo\text{-}cross\text{-}view}}$, which encourages the online encoder to create similar representations for different augmented views of the same graph, and 
(ii) \emph{teacher-student alignment} ~\cite{hinton2015distilling},  $\mathcal{L}_{\mathrm{wo\text{-}cross\text{-}network}}$, which directs the online representations towards more stable targets made by the EMA teacher. Structural regularization is applied to the soft cluster assignments $\mathbf{S}$. Further details are provided in Algorithm \ref{alg:bioc3gnn}.

\subsubsection{Structural Regularization}
\label{sec:struct_reg}
To prevent feature mixing, which leads to oversmoothing, and to obtain better latent representations, we employ a structural regularization based on modularity maximization~\cite{tsitsulin2023graph}, which operates on the soft cluster assignment matrix $\mathbf{S}$. 
Modularity~\cite{tsitsulin2023graph} promotes community-aware clustering by encouraging strong connections within communities and reducing weak connections between them. Let 
$A \in \mathbb{R}^{n \times n}$ 
be the weighted adjacency matrix, 
$d = \sum_j A_{ij}$ 
the degree vector, and 
$m = \sum_{i,j} A_{ij}$ 
the total edge weight. The modularity matrix is defined as
\[
B = A - \frac{d\,d^{\top}}{2m}.
\]
The modularity objective is expressed as
\[
\mathcal{L}_{\mathrm{mod}} = -\frac{1}{2m}\,\mathrm{Tr}(S^{\top} B S).
\]
To avoid collapsing into a single cluster, we add a balancing regularizer
\[
\mathcal{L}_{\mathrm{collapse}} 
= 
\frac{\sqrt{K}}{n}
\left\|
\sum_{i=1}^{n} S_i
\right\|_F
- 1,
\]
where $K$ represents the number of clusters. The final modularity regularization \cite{tsitsulin2023graph} becomes
\[
\mathcal{L}_{\mathrm{struct}}
=
\mathcal{L}_{\mathrm{mod}}
+
\mathcal{L}_{\mathrm{collapse}}.
\]




\subsection{Overall Objective and Optimization} \label{sec:overall}
The final learning goal combines both:
\[
\mathcal{L}
=
\mathcal{L}_{\text{struct}}
+
\lambda_{\text{con}}\,
\mathcal{L}_{\text{con}}.
\]

\begin{figure*}[t]
    \centering
    \includegraphics[
        width=\textwidth,
        height=0.9\textheight,
        keepaspectratio
    ]{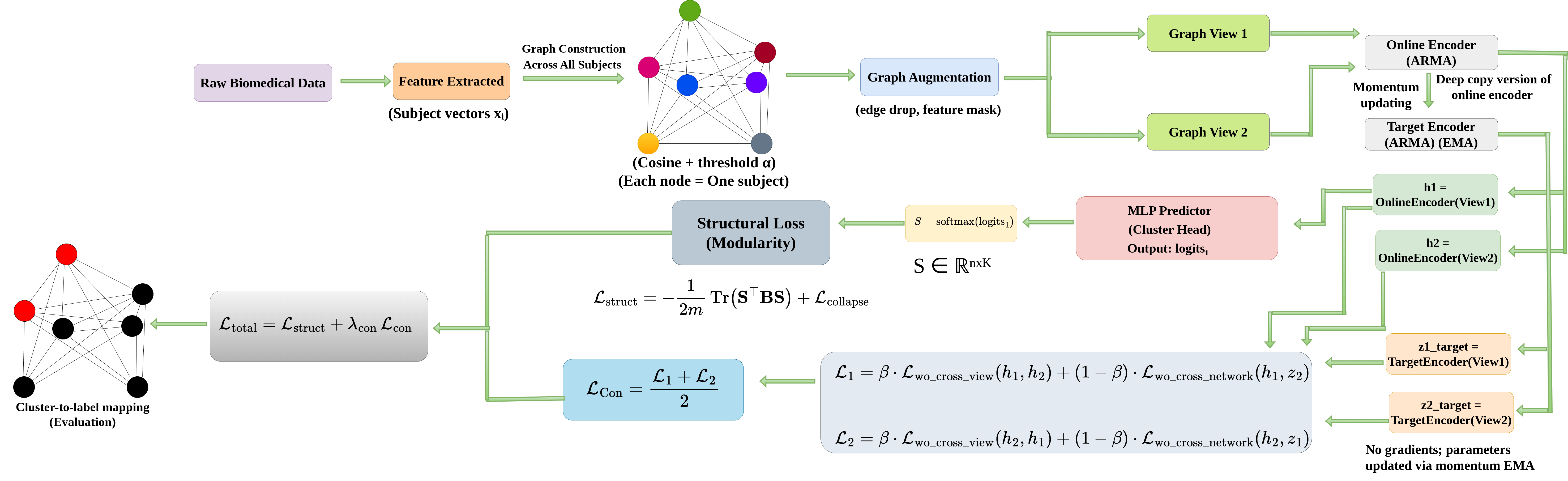}
    \caption{
Overview of the ARMA-C$^3$ framework. A similarity graph is constructed from subject-level features and augmented into multiple views. An ARMA-based encoder learns node representations via joint modularity regularization and contrastive learning.
}
    \label{fig:pipeline}
\end{figure*}

The model is trained using AdamW with an EMA teacher for stability, and the full algorithm is summarized in Algorithm~\ref{alg:bioc3gnn}. An overview of the proposed ARMA-C$^3$ framework is illustrated in Fig.~\ref{fig:pipeline}.
\vspace{0.4cm}
\begin{algorithm}[!ht]
\caption{ARMA-C$^3$: Contrastive ARMA Convolutional Learning}
\label{alg:bioc3gnn}
\small
\begin{algorithmic}[1]
\Require Feature matrix $X \in \mathbb{R}^{n \times d}$,
threshold $\alpha$,
number of clusters $K$,
EMA decay m,
contrastive weight $\lambda_{\mathrm{con}}$
\Ensure Node embeddings and cluster assignments

\State \textbf{Graph Construction:}
$\ell_2$-normalize $X$, compute cosine similarity matrix $W$,
apply threshold $\alpha$ to obtain adjacency $A$,
construct graph $G=(V,E,X,A)$

\State \textbf{Initialization:}
Initialize online encoder $f_{\theta}$,
prediction MLP $g$,
and target encoder $f_{\theta'} \leftarrow f_{\theta}$

\For{each training iteration (mini-batch) $t = 1,\dots,T$}

    \State Sample two graph augmentations $\mathcal{G}_1, \mathcal{G}_2$
    
    \State \textbf{Online encoder:}
    $\mathbf{h}_1 = f_{\theta}(\mathcal{G}_1)$,
    $\mathbf{h}_2 = f_{\theta}(\mathcal{G}_2)$
    
    \State \textbf{Target encoder (EMA, no gradients):}
    $\mathbf{z}_1 = f_{\theta'}(\mathcal{G}_1)$,
    $\mathbf{z}_2 = f_{\theta'}(\mathcal{G}_2)$
    
    \State \textbf{Cluster assignment:}
    $\mathbf{S} = \mathrm{softmax}(g(\mathbf{h}_1)) \in \mathbb{R}^{n \times K}$

    \State \textbf{Structural Loss:}
$
\mathcal{L}_{\mathrm{struct}}
=
-\frac{1}{2m}
\mathrm{Tr}
\left(
\mathbf{S}^{\top}
\left(
A - \frac{\mathbf{d}\mathbf{d}^{\top}}{2m}
\right)
\mathbf{S}
\right)
+ \mathcal{L}_{\mathrm{collapse}}
$
\State \textbf{MERIT Contrastive Loss:}
\begin{align*}
    \mathcal{L}_{\mathrm{wo\text{-}cross\text{-}view}}(\mathbf{h}_1,\mathbf{h}_2)
&= -\frac{1}{n}
\sum_{i=1}^{n}
\log
\frac{
\exp(\mathrm{sim}(\mathbf{h}_{1,i}, \mathbf{h}_{2,i}))
}{
\exp(\mathrm{sim}(\mathbf{h}_{1,i}, \mathbf{h}_{2,i}))
+
\sum_{j \ne i}
\exp(\mathrm{sim}(\mathbf{h}_{1,i}, \mathbf{h}_{1,j}))
}
\\[6pt]
\mathcal{L}_{\mathrm{wo\text{-}cross\text{-}network}}(\mathbf{h}_1,\mathbf{z}_2)
&= -\frac{1}{n}
\sum_{i=1}^{n}
\log
\frac{
\exp(\mathrm{sim}(\mathbf{h}_{1,i}, \mathbf{z}_{2,i}))
}{
\sum_{j=1}^{n}
\exp(\mathrm{sim}(\mathbf{h}_{1,i}, \mathbf{z}_{2,j}))
}
\\[8pt]
\mathcal{L}_1
&= \beta \mathcal{L}_{\mathrm{wo\text{-}cross\text{-}view}}(\mathbf{h}_1,\mathbf{h}_2)
+(1-\beta)\mathcal{L}_{\mathrm{wo\text{-}cross\text{-}network}}(\mathbf{h}_1,\mathbf{z}_2)
\\
\mathcal{L}_2
&= \beta \mathcal{L}_{\mathrm{wo\text{-}cross\text{-}view}}(\mathbf{h}_2,\mathbf{h}_1)
+(1-\beta)\mathcal{L}_{\mathrm{wo\text{-}cross\text{-}network}}(\mathbf{h}_2,\mathbf{z}_1)
\\
\mathcal{L}_{\mathrm{con}}
&= \frac{\mathcal{L}_1+\mathcal{L}_2}{2} , \text{Where } \mathrm{sim}(\cdot,\cdot) \text{ denotes cosine similarity.}
\end{align*}

    \State \textbf{Total loss and update:}
    \[
    \theta \leftarrow
    \arg\min_{\theta}
    \left(
    \mathcal{L}_{\mathrm{struct}}
    +
    \lambda_{\mathrm{con}} \mathcal{L}_{\mathrm{con}}
    \right)
    \]

    \State \textbf{EMA update:}
    \[
    \theta' \leftarrow m \theta' + (1-m)\theta , \text{Where } m \text{ is momentum parameter.}
    \]

\EndFor

\State \textbf{Output:}
Final node embeddings and cluster labels from $f_{\theta}$
\end{algorithmic}
\end{algorithm}

\subsection{Graph-based Semi-supervised Learning}  
\label{sec:transductive}  
ARMA-C$^3$ also operates in a graph-based semi-supervised setting. Here, only some subjects are labeled, while the graph structure helps with representation learning and node classification. This approach allows the framework to take advantage of graph structure, feature similarity, and limited supervision to improve predictive performance. When labels are available, the parameters are updated using $\mathcal{L}_{\mathrm{sup}}$ in this setup, which mirrors real biomedical situations, where labeled samples are hard to find and expensive to acquire.

Let $\mathcal{V}_L$ and $\mathcal{V}_U$ represent the labeled and unlabeled nodes, respectively. We use a cross-entropy loss  
\[
\mathcal{L}_{\mathrm{sup}}  
=  
-\sum_{v_i\in\mathcal{V}_L}  
\sum_{j=1}^{n}\mathbf{y}_{ij}\log(\hat{\mathbf{y}}_{ij})  
\]
along with contrastive and structural regularization terms. This results in the total optimization objective:
\[
\mathcal{L}_{\mathrm{total}}
=
\lambda_{con}\,\mathcal{L}_{\mathrm{con}}
+
\mathcal{L}_{\mathrm{sup}}
+
\lambda_{struct}\,\mathcal{L}_{\mathrm{struct}}.
\]
By passing information through graph connections, both labeled and unlabeled subjects work together in representation learning. This helps the framework capture relational and topological links between subjects, which enhances robustness and generalization with limited supervision. The proposed semi-supervised framework operates in a transductive setting, where the graph is constructed using all subjects prior to optimization, while supervision is applied only to labeled nodes. Although this setup requires graph reconstruction when new subjects are introduced, it remains appropriate for cohort-level biomedical analysis scenarios, where the objective is to jointly model relational structure and disease heterogeneity within a fixed study population.

\section{Experiments, Results and Discussion}
\label{sec:exp_res_disc}
This section evaluates the proposed ARMA-C$^3$ framework across multiple
biomedical imaging modalities to assess its effectiveness, robustness, and
generalization capability. We report quantitative results, qualitative
analyses, and ablation studies to analyze the contribution of each model
component, followed by a discussion of key observations and practical
implications.

\subsection{Experimental Setup}
ARMA-C$^3$ is evaluated on five biomedical datasets spanning diffusion MRI,
ultrasound, and chest X-ray imaging. Subject-level feature representations
obtained in Section~\ref{Method} are used to construct subject graphs for all
experiments. A unified training and evaluation protocol is adopted to ensure
consistent and modality-independent comparison.

\subsubsection{Datasets}
We used 390 subjects from ADNI (CN:133, MCI:167, AD:90) with diffusion MRI acquired using Siemens scanners with 46, 54, 55 gradient directions. 
NIFD  comprises 146 subjects (FTD: 98, controls: 48) with high-angular resolution diffusion MRI (120 gradient directions) for neurodegenerative evaluation.
BreastMNIST contains 780 ultrasound images (benign: 210, malignant: 570) for classification with class imbalance. 
PneumoniaMNIST provides 3,583 pediatric chest X-rays (normal: 2000, pneumonia: 1583) for radiological disease screening. The  liver Ultrasound dataset~\cite{xu2022annotated} contains 635 ultrasound images (benign: 200, malignant: 435) for liver ultrasound classification with a noticeable class imbalance. All available image samples were used for experimentation.

\begin{figure}[!htbp]
\centering
\captionsetup[subfigure]{justification=centering}

\begin{subfigure}[t]{0.18\textwidth}
\centering
\resizebox{\linewidth}{3.0cm}{%
\includegraphics[trim={2.0cm 0.5cm 2.0cm 0.5cm}, clip]{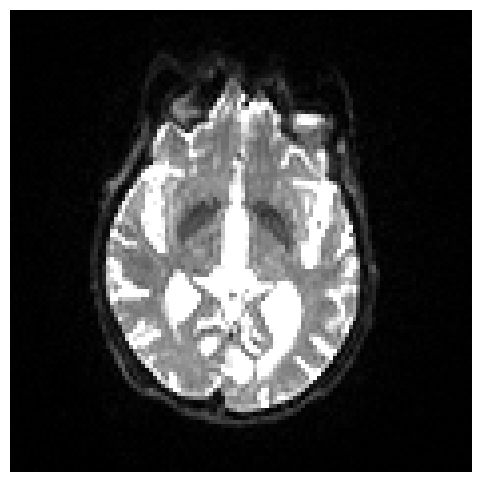}%
}
\caption{CN}
\end{subfigure}\hfill
\begin{subfigure}[t]{0.18\textwidth}
\centering
\resizebox{\linewidth}{3.0cm}{%
\includegraphics[trim={2.0cm 0.5cm 2.0cm 0.5cm}, clip]{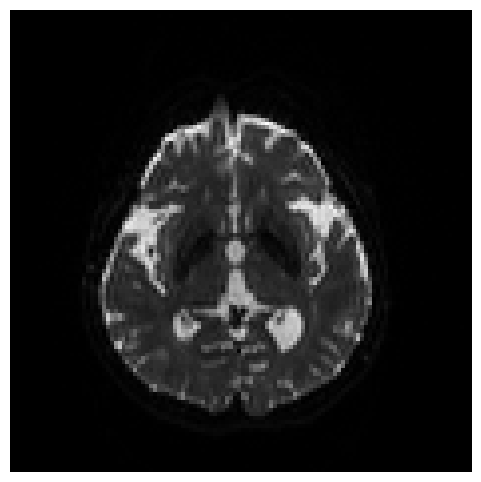}%
}
\caption{MCI}
\end{subfigure}\hfill
\begin{subfigure}[t]{0.18\textwidth}
\centering
\resizebox{\linewidth}{3.0cm}{%
\includegraphics[trim={2.0cm 0.5cm 2.0cm 0.5cm}, clip]{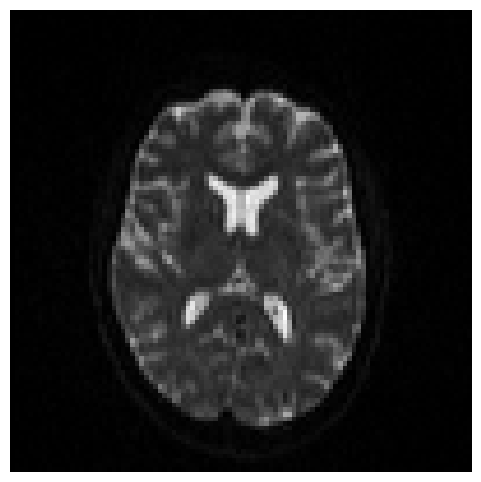}%
}
\caption{NIFD Ctrl.}
\end{subfigure}\hfill
\begin{subfigure}[t]{0.18\textwidth}
\centering
\resizebox{\linewidth}{3.0cm}{%
\includegraphics[trim={2.0cm 0.5cm 2.0cm 0.5cm}, clip]{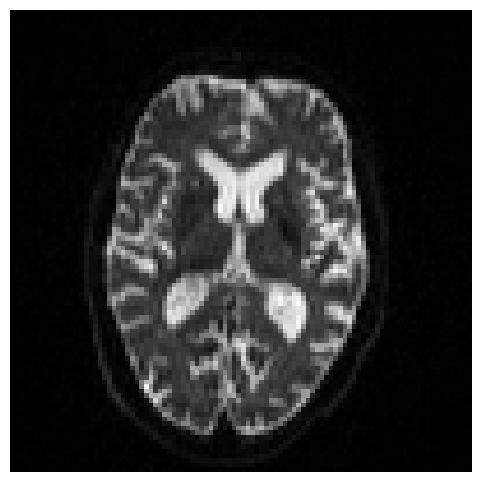}%
}
\caption{NIFD Pat.}
\end{subfigure}\hfill
\begin{subfigure}[t]{0.18\textwidth}
\centering
\resizebox{\linewidth}{3.0cm}{%
\includegraphics[trim={0.5cm 0.5cm 0.5cm 0.5cm}, clip]{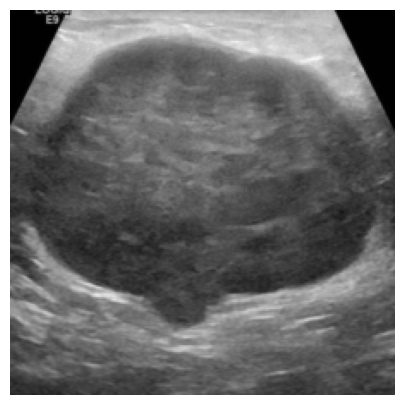}%
}
\caption{Breast Ben.}
\end{subfigure}

\vspace{0.35cm}

\begin{subfigure}[t]{0.18\textwidth}
\centering
\resizebox{\linewidth}{3.0cm}{%
\includegraphics[trim={0cm 0.5cm 0cm 0.5cm}, clip]{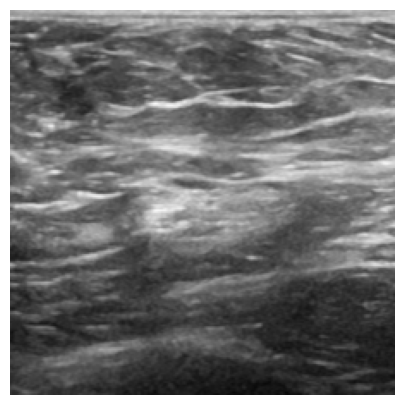}%
}
\caption{Breast Malig.}
\end{subfigure}\hfill
\begin{subfigure}[t]{0.18\textwidth}
\centering
\resizebox{\linewidth}{3.0cm}{%
\includegraphics[trim={2.0cm 0.5cm 2.0cm 0.5cm}, clip]{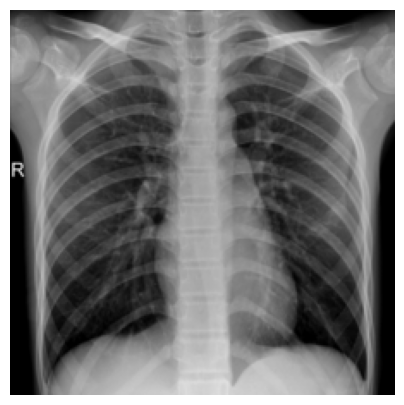}%
}
\caption{Chest Norm.}
\end{subfigure}\hfill
\begin{subfigure}[t]{0.18\textwidth}
\centering
\resizebox{\linewidth}{3.0cm}{%
\includegraphics[trim={2.0cm 0.5cm 2.0cm 0.5cm}, clip]{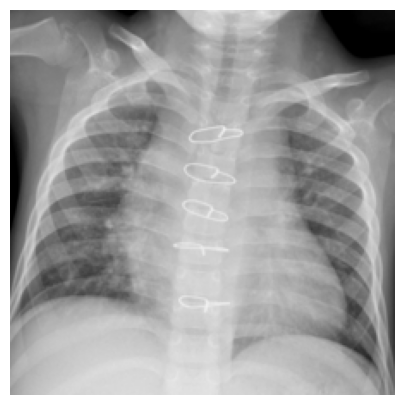}%
}
\caption{Chest Pneu.}
\end{subfigure}\hfill
\begin{subfigure}[t]{0.18\textwidth}
\centering
\resizebox{\linewidth}{3.0cm}{%
\includegraphics[trim={2.0cm 0.5cm 2.0cm 0.5cm}, clip]{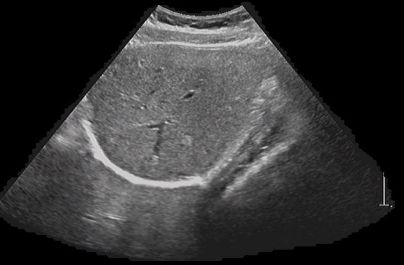}
}
\caption{Liver Ben.}
\end{subfigure}\hfill
\begin{subfigure}[t]{0.18\textwidth}
\centering
\resizebox{\linewidth}{3.0cm}{%
\includegraphics[trim={2.0cm 0.5cm 2.0cm 0.5cm}, clip]{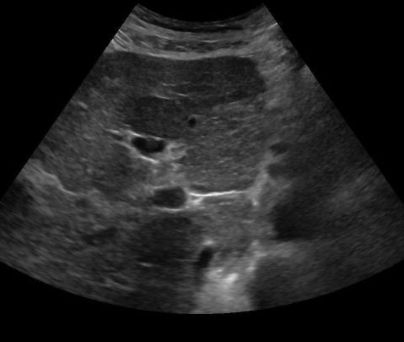}

}
\caption{Liver Malig.}
\end{subfigure}

\caption{Representative examples from the biomedical datasets used in this study, including diffusion MRI slices from ADNI and NIFD cohorts, breast ultrasound images, chest X-ray images, and liver ultrasound images.}
\label{fig:dataset_examples}
\end{figure}

\subsubsection{Implementation Details \& Hyperparameters}

We evaluate ARMA-C$^3$ using diffusion MRI data from cognitively normal (CN), mild cognitive impairment (MCI), and Alzheimer’s disease (AD) subjects. Our analysis focuses on white-matter pathways known to exhibit early microstructural degeneration during AD progression, as reflected by reductions in Fractional Anisotropy (FA). 
The selected regions of interest (ROIs) include the corpus callosum, fornix, right corticospinal tract, right cerebral peduncle, right anterior limb of the internal capsule, and the cingulum bundle. These regions are chosen based on strong neuroanatomical evidence of their vulnerability to Alzheimer’s disease (AD) pathology. The corpus callosum supports inter-hemispheric cognitive integration and exhibits fractional anisotropy (FA) decline due to axonal demyelination \cite{di2010vivo, walterfang2014shape, hua2008tract}. The fornix, a primary hippocampal efferent pathway, shows FA reduction that precedes hippocampal atrophy and correlates with episodic memory impairment \cite{oishi2011multi, nowrangi2015fornix, kantarci2014white}. The right corticospinal tract demonstrates FA decreases associated with disrupted axonal organization and subtle motor deficits \cite{teipel2014fractional}, while the right cerebral peduncle shows FA alterations linked to executive dysfunction \cite{douaud2011dti}. The right anterior limb of the internal capsule is associated with impairments in processing speed and working memory, often exhibiting hemispheric asymmetry \cite{madden2008cerebral, mielke2009regionally}. Finally, the cingulum bundle, particularly its hippocampal subdivision, exhibits early FA reductions correlated with episodic memory decline and progressive degeneration across its segments \cite{villain2010sequential, zhuang2010white, zhang2007diffusion, yasmin2009tract}.

For all subjects, FA maps are computed using the DIPY library~\cite{garyfallidis2014dipy}. Each scan is registered to a standard white-matter atlas, and FA distributions were extracted from $p = 9$ neuroanatomically meaningful ROIs. For each ROI, we computed normalized FA histograms using $q = 20$ bins, yielding compact $180$-dimensional subject-level descriptors.

These features are used as node attributes in the graph and processed using an ARMA-based graph encoder. Unless otherwise specified, ARMA-C$^3$ employs a single ARMA layer with (\texttt{num\_stacks}=1, \texttt{num\_layers}=1), followed by batch normalization, dropout, and a linear projection head. The NIFD dataset benefited from deeper message propagation and therefore utilized a deeper ARMA encoder with (\texttt{num\_stacks}=1, \texttt{num\_layers}=3) and a larger hidden dimensionality.

All models are optimized using AdamW with a learning rate of $1\times10^{-4}$ and weight decay of $1\times10^{-4}$. A StepLR scheduler with a step size of 200 epochs and a decay factor of 0.5 was applied to stabilize training over long horizons. All experiments were conducted on an NVIDIA RTX A4000 GPU. For a single training run, ARMA-C$^3$ required approximately 0.64 min (61.46 MB GPU memory) for CN vs. MCI (5000 epochs), 0.72 min (119.03 MB) for CN vs. AD (5000 epochs), 1.23 min (109.12 MB) for NIFD (5000 epochs), 0.91 min (116.28 MB) for the liver ultrasound dataset (2500 epochs), 3.27 min (889.78 MB) for BreastMNIST (2500 epochs), and 8.61 min (2977.96 MB) for PneumoniaMNIST (2500 epochs), demonstrating practical computational efficiency across datasets of varying scale and modality complexity. 

To reduce sensitivity to label sampling variability, particularly for small cohorts such as NIFD, all semi-supervised results are averaged over 20 repeated folds with different random seeds for labeled subject selection.

Dataset-specific hyperparameters, including the graph sparsification threshold $\alpha$, contrastive loss coefficient $\lambda_{\text{con}}$, activation function, hidden feature dimension, and number of training epochs, are summarized in Table~\ref{tab:unified_hyperparams} and are selected via grid search.\\
We have performed an extensive systematic ablation study across various architectural and hyperparameter choices as shown in Table~\ref{tab:unified_hyperparams} and report the corresponding performance obtained under the selected hyperparameter configuration.

\noindent\textbf{Computational Complexity and Scalability:}
Graph construction requires $O(E d)$ operations, while modularity computation incurs an $O(E)$ cost. Where $E$ denotes the number of graph edges and $d$ denotes the dimensionality of the node feature vectors. To improve scalability, we apply threshold-based sparsification controlled by $\alpha$. For larger graphs (e.g., PneumoniaMNIST), a high threshold ($\alpha = 0.9$) significantly reduces edge density, lowering memory usage and message passing complexity. Increasing $\alpha$ makes the graph sparser and computation more efficient, but may remove weaker connections, introducing a trade-off between efficiency and structural information preservation.

\begin{table}[!htbp]
\centering
\caption{Hyperparameters for ARMA-C$^3$. semi-supervised uses a 10\% train / 90\% test split.}
\label{tab:unified_hyperparams}
\setlength{\tabcolsep}{5pt}
\renewcommand{\arraystretch}{0.9}
\begin{tabular}{lcccccc}
\toprule
\textbf{Set} & \textbf{Dataset} & $\alpha$ & $\lambda_{\text{con}}$ & Dim & Act & Epochs \\
\midrule
\multirow{5}{*}{Unsup}
& CN--MCI & 0.92 & $3\times10^{-1}$ & 256 & ELU  & 5000 \\
& CN--AD  & 0.80 & $5\times10^{-1}$ & 256 & ELU  & 5000 \\
& NIFD    & 0.50 & $10^{-2}$  & 512 & ELU  & 5000 \\
& Breast  & 0.73 & $10^{-3}$        & 256 & ReLU & 2500 \\
& Pneumo  & 0.90 & $5\times10^{0}$ & 256 & ELU  & 2500 \\
& Liver & 0.83 & 8 & 256 & SiLU & 2500 \\
\midrule
\multirow{5}{*}{Semi}
& CN--MCI & 0.92 & $10^{-4}$        & 256 & ReLU & 2000 \\
& CN--AD  & 0.80 & $10^{-1}$        & 256 & ELU  & 2000 \\
& NIFD    & 0.50 & $10^{-2}$        & 256 & SELU & 2000 \\
& Breast  & 0.73 & $10^{-4}$        & 256 & ReLU & 2000 \\
& Pneumo  & 0.90 & $5\times10^{-3}$ & 256 & ELU  & 2000 \\
& Liver & 0.83 & $10^{-3}$ & 256 & SiLU & 2500 \\
\bottomrule
\end{tabular}
\end{table}

\subsection{Results}

\paragraph{Unsupervised Setting Observations:}
Across all datasets, we evaluate multiple ablation settings using Modularity~\cite{tsitsulin2023graph} and Normalized Cut~\cite{bianchi2020spectral} as structural regularization losses ($\mathcal{L}_{\mathrm{struct}}$), together with DGI~\cite{velickovic2019deep} and MERIT~\cite{jin2021multi} as contrastive learning objectives ($\mathcal{L}_{\mathrm{con}}$), reported under the \textit{Loss Function} categories in the corresponding results tables.\\Here, \textit{Mod} in the tables denotes Modularity-based structural regularization.
For the challenging CN vs. MCI task (Table~\ref{tab:combined_unsup_results}(a)), classical clustering methods show limited discriminative ability. Graph-based models generally improve clustering performance, with ARMA-C$^3$ variants demonstrating competitive overall results across evaluation metrics. In particular, ARMA-C$^3$ and the GCN backbone encoder with cut-based structural regularization loss ($\mathcal{L}_{\mathrm{struct}}$) and MERIT contrastive learning loss ($\mathcal{L}_{\mathrm{con}}$) obtain the highest accuracy and F1-scores, indicating that regularization and augmentation-invariant representations are crucial when disease boundaries are subtle. \\
In the CN vs. AD setting (Table~\ref{tab:combined_unsup_results}(b)), the GAT backbone encoder with Modularity-based structural regularization loss ($\mathcal{L}_{\mathrm{struct}}$) achieves the highest accuracy (0.785), marginally outperforming ARMA-C$^3$ (0.748). This behavior can be attributed to the clearer class separation between CN and AD subjects, where disease-driven structural differences are more pronounced and attention-based neighborhood aggregation can effectively exploit strong local community boundaries without requiring augmentation-invariant consistency. In contrast, ARMA-C$^3$ achieves the highest recall, indicating improved sensitivity toward detecting true AD subjects, which is clinically important in disease screening settings. Furthermore, the ARMA backbone encoder with Modularity-based structural regularization and DGI contrastive learning achieves the best F1-score, suggesting an improved balance between cluster purity and completeness.

On the NIFD dataset (Table~\ref{tab:nifd_combined}), ARMA-C$^3$ achieves competitive performance relative to traditional clustering approaches and graph-based baselines, attaining the highest accuracy across evaluated methods.
For BreastMNIST (Table~\ref{tab:combined_medmnist_unsup}(a)), ARMA-C$^3$ achieves strong performance, particularly in Recall and F1-score, indicating effective sensitivity toward malignant samples under class imbalance, while reflecting the expected precision--recall trade-off. On PneumoniaMNIST (Table~\ref{tab:combined_medmnist_unsup}(b)), the GCN backbone encoder with modularity-based structural regularization ($\mathcal{L}_{\mathrm{struct}}$) and MERIT contrastive learning ($\mathcal{L}_{\mathrm{con}}$) achieves the highest accuracy and F1-score, while ARMA-C$^3$ also demonstrates competitive performance on radiological datasets with clearer structural patterns. On the liver ultrasound dataset (Table~\ref{tab:liver_unsup_results}(a)), ARMA-C$^3$ achieves strong clustering performance, demonstrating robust separation between benign and malignant liver ultrasound cases under class imbalance.
Several baseline methods on highly imbalanced cohorts (e.g., the GAT backbone encoder with modularity-based structural regularization and MERIT contrastive learning on NIFD) exhibit near-majority-class prediction behaviour, producing inflated recall despite limited minority-class discrimination. In contrast, ARMA-C$^3$ maintains a more balanced precision--recall trade-off across diffusion-MRI and radiological datasets.
For statistical testing, results are reported using the exact mean and standard deviation computed across 10 independent runs prior to rounding for tabular presentation.
To further evaluate statistical robustness, we performed Wilcoxon signed-rank tests comparing ARMA-C$^3$ against MAGI~\cite{liu2024revisiting}, a recent graph clustering framework that reformulates modularity maximization through contrastive learning, treating community structure as the source of positive and negative pairs. Across 10 independent runs, ARMA-C$^3$ consistently achieved statistically significant improvements over MAGI on all evaluated datasets. Specifically, ARMA-C$^3$ achieved higher mean F1-scores on CN vs.\ MCI ($0.786 \pm 0.004$ vs.\ $0.778 \pm 0.005$, $p = 0.0049$), CN vs.\ AD ($0.711 \pm 0.005$ vs.\ $0.340 \pm 0.032$, $p = 0.0010$), NIFD ($0.743 \pm 0.008$ vs.\ $0.328 \pm 0.030$, $p = 0.0010$), BreastMNIST ($0.841 \pm 0.010$ vs.\ $0.790 \pm 0.001$, $p = 0.0010$), PneumoniaMNIST ($0.915 \pm 0.000$ vs.\ $0.906 \pm 0.001$, $p = 0.0010$), and liver ultrasound ($0.801 \pm 0.008$ vs.\ $0.634 \pm 0.089$, $p = 0.0010$). These findings suggest that ARMA-C$^3$ learns more discriminative graph representations across diverse biomedical clustering tasks.

To assess the robustness of the learned feature representations, we evaluate the effect of dimensionality reduction via PCA followed by K-Means clustering. K-Means is applied both on the original feature space and on the PCA-projected features (2D) to examine how well the underlying cluster structure is preserved under projection. For standard K-Means operating on raw features, clustering performance degrades after PCA projection, revealing sensitivity to the feature space and limited intrinsic separability. This degradation can be attributed to the absence of adaptive spectral regularization: without learnable graph filters, raw feature representations retain high-frequency spectral components associated with noise and intra-class variance, which corrupt the principal directions estimated by PCA and misalign the projected features with the true cluster manifolds. In contrast, applying the same PCA + K-Means procedure to embeddings produced by the proposed graph-based model yields stable clustering performance, demonstrating that the learned representations preserve the intrinsic data geometry even under aggressive dimensionality reduction. This robustness may be attributed to the learnable spectral filters, which suppress task-irrelevant high-frequency components and concentrate discriminative variance along directions that remain coherent after projection. The corresponding PCA + K-Means results obtained using embeddings learned by the proposed framework are reported in the tables under the \textit{Embedding} model category. 

Furthermore, ARMA-C$^3$ achieves competitive clustering performance relative to the modularity-based variant, indicating enhanced inter-cluster separability and improved representational quality in the learned embedding space.

\begin{figure*}[!htbp]
\centering
\begin{subfigure}{0.26\textwidth}
\centering
\includegraphics[width=\linewidth]{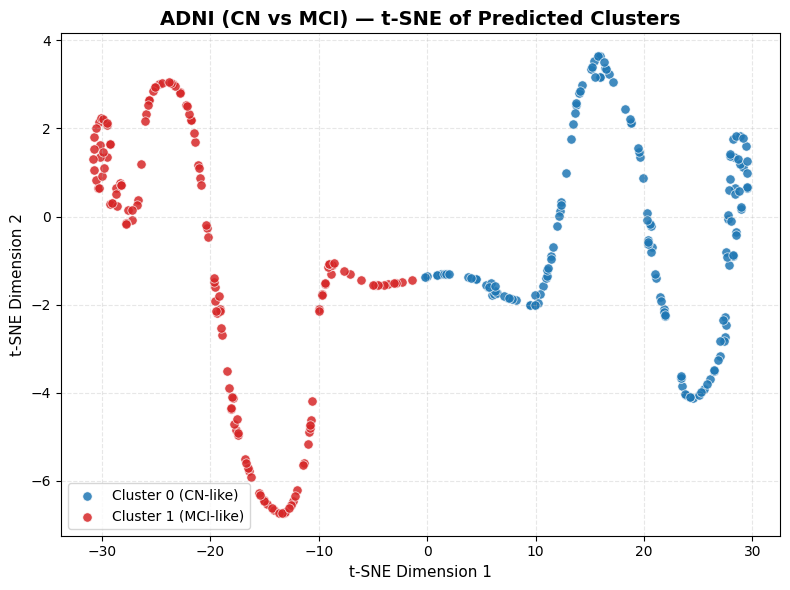}
\caption{CN vs.\ MCI}
\end{subfigure}
\hspace{0.03\textwidth}
\begin{subfigure}{0.26\textwidth}
\centering
\includegraphics[width=\linewidth]{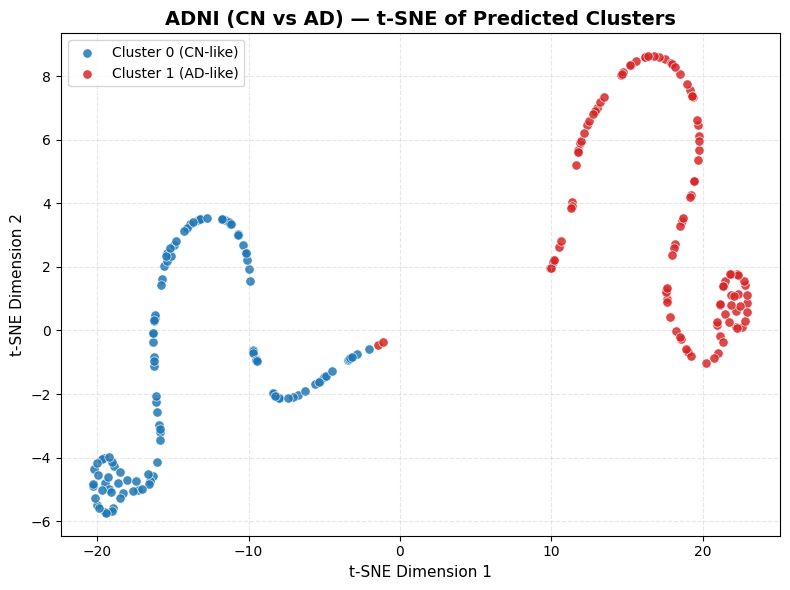}
\caption{CN vs.\ AD}
\end{subfigure}
\hspace{0.03\textwidth}
\begin{subfigure}{0.26\textwidth}
\centering
\includegraphics[width=\linewidth]{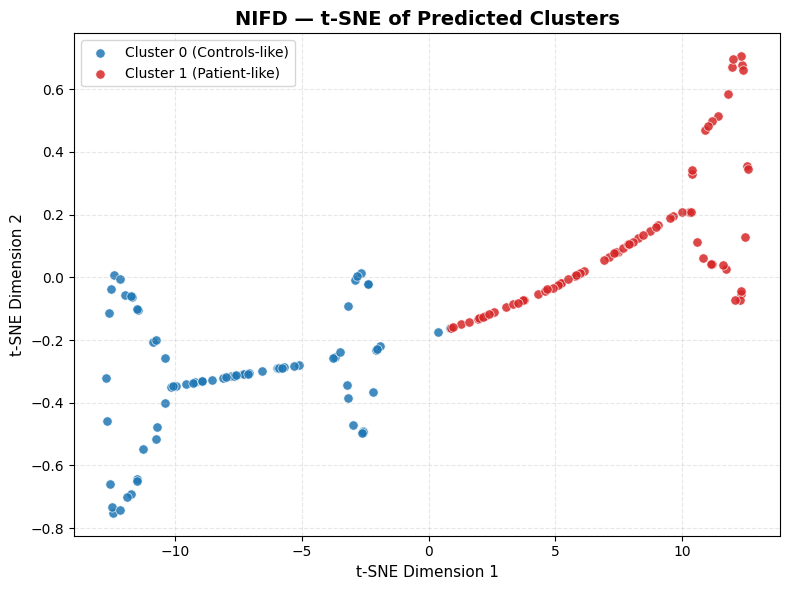}
\caption{NIFD}
\end{subfigure}

\vspace{0.4cm}

\begin{subfigure}{0.26\textwidth}
\centering
\includegraphics[width=\linewidth]{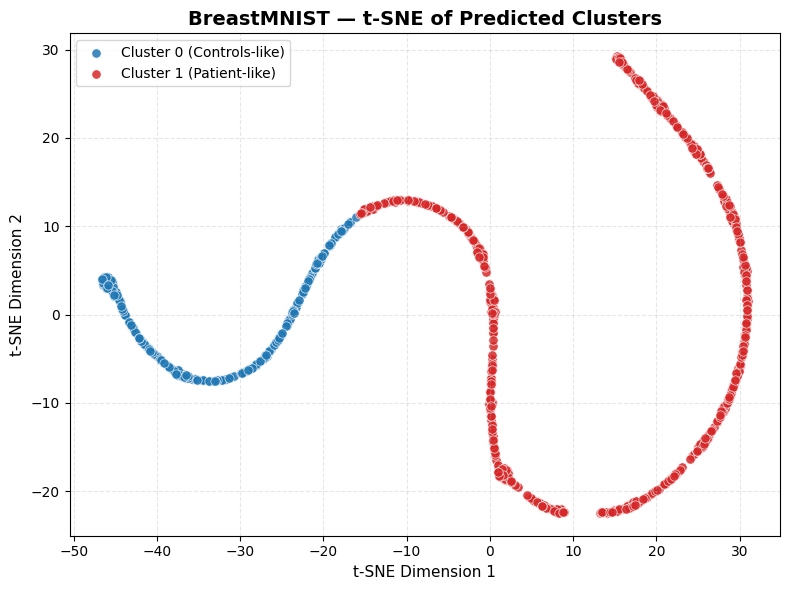}
\caption{BreastMNIST}
\end{subfigure}
\hspace{0.03\textwidth}
\begin{subfigure}{0.26\textwidth}
\centering
\includegraphics[width=\linewidth]{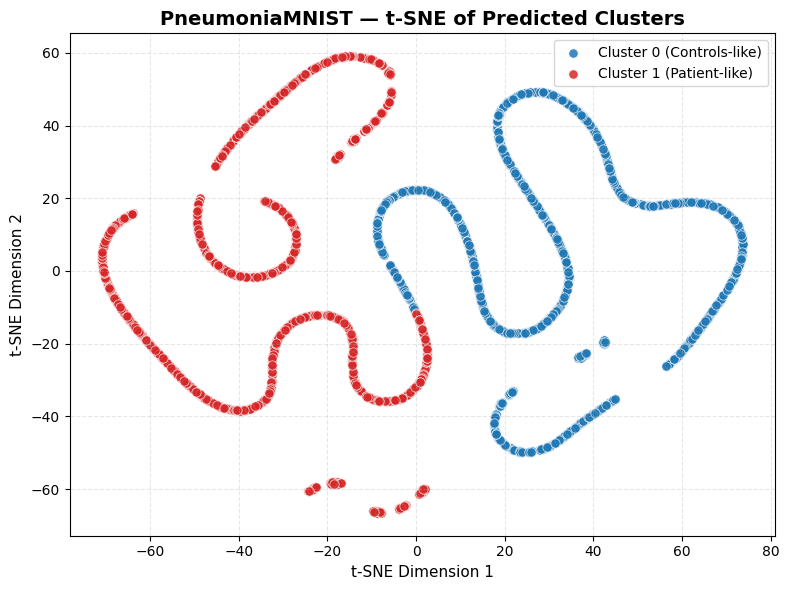}
\caption{PneumoniaMNIST}
\end{subfigure}
\hspace{0.03\textwidth}
\begin{subfigure}{0.26\textwidth}
\centering
\includegraphics[width=\linewidth]{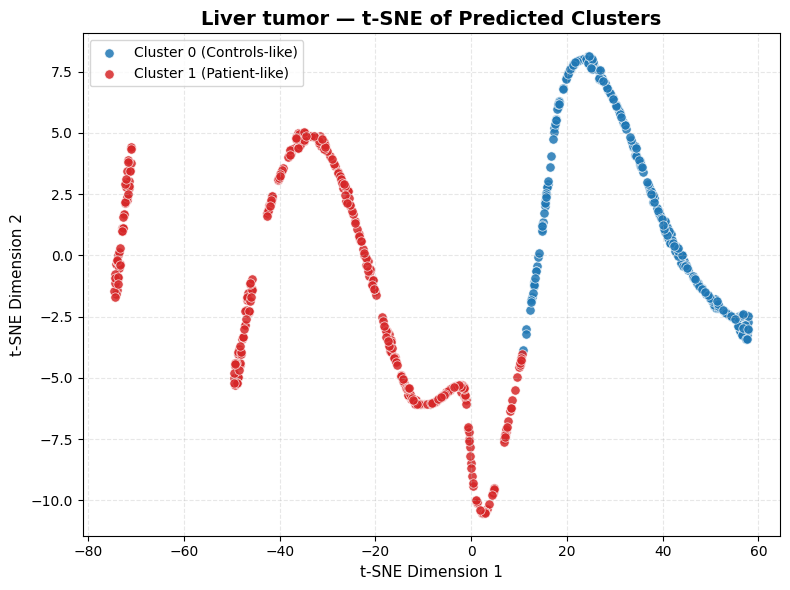}
\caption{Liver Ultrasound}
\end{subfigure}

\caption{t-SNE visualizations of unsupervised node embeddings learned by ARMA-C$^3$ across datasets.}
\label{fig:tsne_unsupervised}
\end{figure*}

This is further confirmed by t-SNE visualizations in Figure \ref{fig:tsne_unsupervised} of the learned feature embeddings $\textbf{h}_1$, which show improved class separations for all datasets.
\begin{figure*}[!htbp]
\centering
\begin{subfigure}{0.25\textwidth}
\centering
\includegraphics[width=\linewidth]{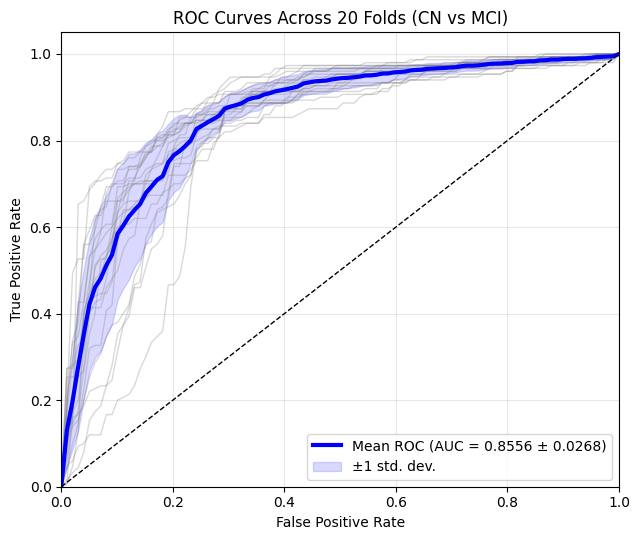}
\caption{CN vs.\ MCI}
\end{subfigure}
\hspace{0.03\textwidth}
\begin{subfigure}{0.25\textwidth}
\centering
\includegraphics[width=\linewidth]{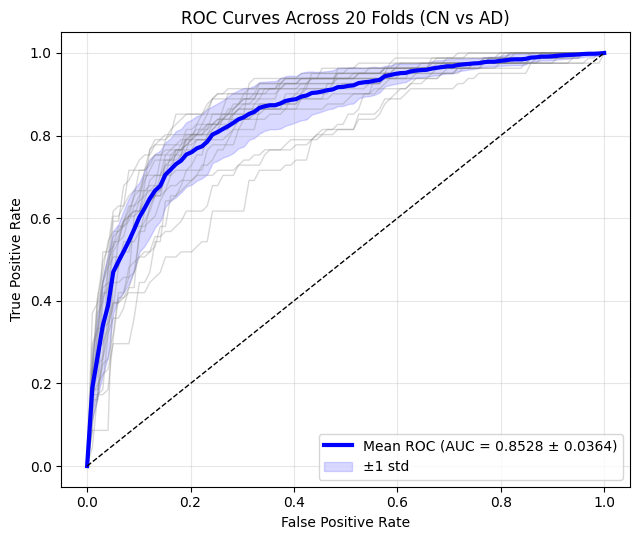}
\caption{CN vs.\ AD}
\end{subfigure}
\hspace{0.03\textwidth}
\begin{subfigure}{0.25\textwidth}
\centering
\includegraphics[width=\linewidth]{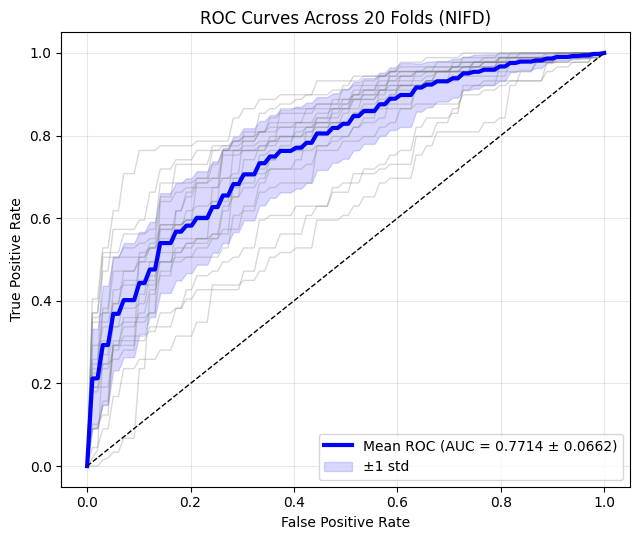}
\caption{NIFD}
\end{subfigure}

\vspace{0.4cm}

\begin{subfigure}{0.25\textwidth}
\centering
\includegraphics[width=\linewidth]{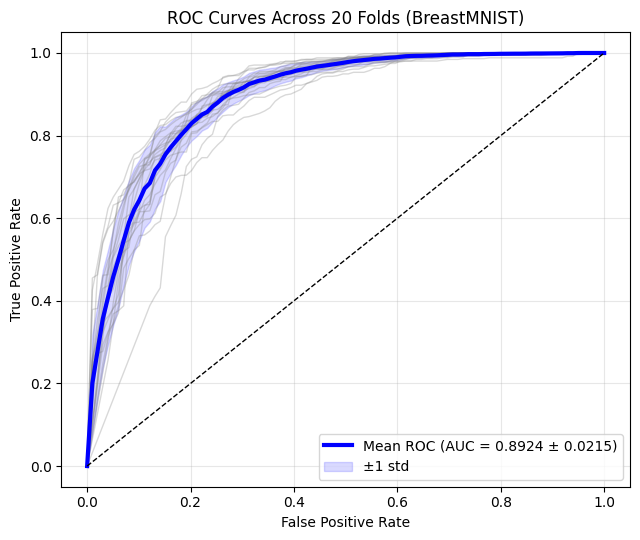}
\caption{BreastMNIST}
\end{subfigure}
\hspace{0.04\textwidth}
\begin{subfigure}{0.25\textwidth}
\centering
\includegraphics[width=\linewidth]{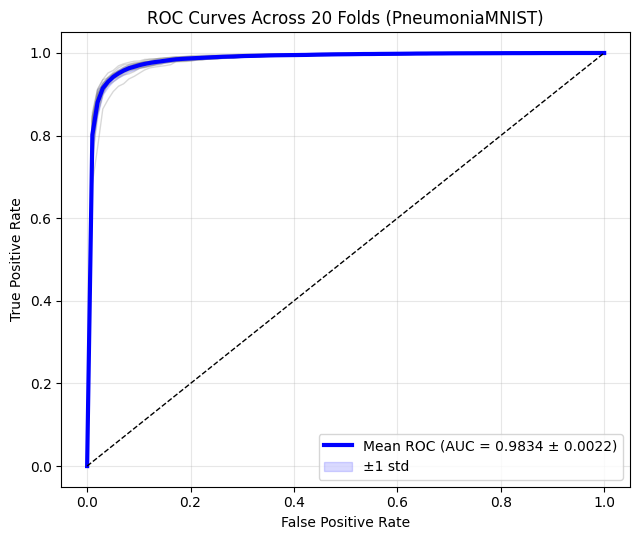}
\caption{PneumoniaMNIST}
\end{subfigure}
\hspace{0.04\textwidth}
\begin{subfigure}{0.25\textwidth}
\centering
\includegraphics[width=\linewidth]{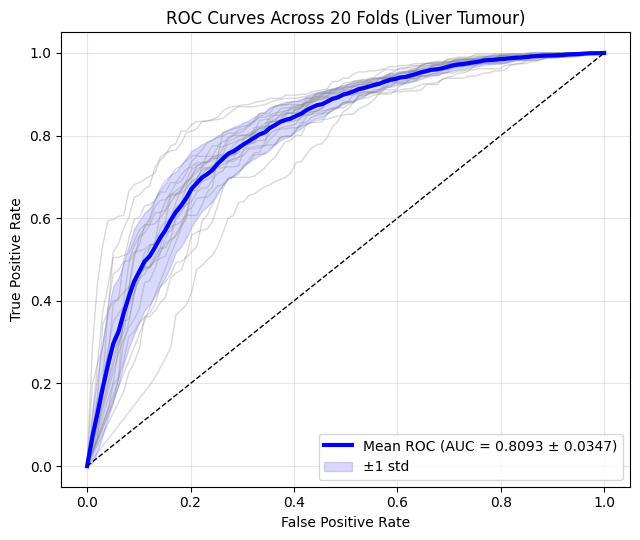}
\caption{Liver Ultrasound}
\end{subfigure}

\caption{ROC curves for semi-supervised classification using ARMA-C$^3$ across datasets.}
\label{fig:roc_semisupervised}
\end{figure*}
\paragraph{Semi-supervised setting observation:}
To assess whether ARMA-C$^3$ can effectively leverage limited supervision, we evaluate a semi-supervised setup using only 10\% labeled data. The structural regularization coefficient $\lambda_{\text{struct}}$ is fixed to 1 throughout all experiments. The learned representations demonstrate robust classification performance across datasets, further supported by the ROC curves in Figure~\ref{fig:roc_semisupervised}.

On the NIFD dataset (Table~\ref{tab:nifd_combined}), semi-supervision improves overall stability across models. While classical classifiers achieve high recall, ARMA-C$^3$ variants provide more balanced performance. 
For ADNI, semi-supervised learning consistently improves classification performance. In the CN vs. MCI task (Table~\ref{tab:adni_combined}(a)), ARMA backbone encoders with Modularity/Cut-based structural regularization ($\mathcal{L}_{\mathrm{struct}}$) achieve strong accuracy, while ARMA-C$^3$ attains the best recall score. In the CN vs.\ AD setting (Table~\ref{tab:adni_combined}(b)), the GAT backbone encoder with Modularity-based structural regularization achieves the highest accuracy (0.791), marginally outperforming ARMA-C$^3$ (0.779). ARMA Cut achieves the highest recall (0.834), which is important in disease-screening scenarios where false negatives carry higher clinical cost, while ARMA-C$^3$ maintains balanced precision--recall behaviour. These results suggest that attention-based methods are particularly effective when disease boundaries are pronounced, whereas ARMA-C$^3$ provides more consistent trade-offs across imbalanced cohorts.
On the BreastMNIST dataset (Table~\ref{tab:medmnist_combined}(a)), ARMA-C$^3$ achieves the highest accuracy and recall, while the ARMA backbone encoder with MERIT contrastive learning ($\mathcal{L}_{\mathrm{con}}$) and Cut-based structural regularization ($\mathcal{L}_{\mathrm{struct}}$) attains the best F1-score. On PneumoniaMNIST (Table~\ref{tab:medmnist_combined}(b)), ARMA-C$^3$ demonstrates strong performance across multiple metrics, obtaining the highest accuracy, recall, and competitive F1-scores, highlighting the effectiveness of community-aware contrastive graph learning on radiological datasets.

\begin{table*}[!htbp]
\centering
\caption{Comparison of unsupervised clustering performance on ADNI cohorts using traditional and graph-based approaches.
Metrics are reported as mean $\pm$ standard deviation over ten repeated experiments.}
\label{tab:combined_unsup_results}

\begin{subtable}{0.78\textwidth}
\centering
\caption{CN vs MCI}
\setlength\tabcolsep{5pt}
\begin{adjustbox}{max width=\textwidth}
{\large
\begin{tabular}{@{} l l c c c c c @{}}
\toprule
\textbf{Model} & \textbf{Loss Function} & $\boldsymbol{\lambda_{con}}$ 
& \textbf{Accuracy} & \textbf{Precision} & \textbf{Recall} & \textbf{F1 Score} \\
\midrule
K-Means ~\cite{macqueen1967kmeans} & -- & -- 
& 0.721 $\pm$ 0.01 & 0.804 $\pm$ 0.01 & 0.660 $\pm$ 0.02 & 0.724 $\pm$ 0.01 \\
PCA + K-Means & -- & -- 
& 0.697 $\pm$ 0.05 
& 0.774 $\pm$ 0.07 
& 0.684 $\pm$ 0.10 
& 0.716 $\pm$ 0.01 \\
Token-Cut \cite{wang2023tokencut} & -- & -- 
& 0.713 $\pm$ 0.00 & 0.776 $\pm$ 0.00 & 0.683 $\pm$ 0.00 & 0.726 $\pm$ 0.00 \\
SpectralNet \cite{shaham2018spectralnet} & -- & -- 
& 0.739 $\pm$ 0.02 & 0.622 $\pm$ 0.24 & 0.544 $\pm$ 0.19 & 0.580 $\pm$ 0.21 \\
TANGO \cite{ma2024tango} & -- & -- 
& 0.617 $\pm$ 0.00 & 0.596 $\pm$ 0.00 & \textbf{0.970 $\pm$ 0.00} & 0.738 $\pm$ 0.00 \\
\textbf{MAGI \cite{liu2024revisiting}} & -- & -- 
& \textbf{0.774 $\pm$ 0.00} & \textbf{0.863 $\pm$ 0.02} & 0.706 $\pm$ 0.02 & 0.776 $\pm$ 0.01 \\
\midrule
GCN~\cite{kipf2017semi} & Mod~\cite{tsitsulin2023graph} & N/A 
& 0.750 $\pm$ 0.00 & 0.804 $\pm$ 0.00 & 0.730 $\pm$ 0.00 & 0.765 $\pm$ 0.00 \\

GAT~\cite{velickovic2018graph} & Mod & N/A 
& 0.757 $\pm$ 0.01 & 0.813 $\pm$ 0.01 & 0.731 $\pm$ 0.00 & 0.770 $\pm$ 0.01 \\

ARMA~\cite{bianchi2021graph} & Mod & N/A 
& 0.749 $\pm$ 0.01 & 0.805 $\pm$ 0.01 & 0.725 $\pm$ 0.00 & 0.763 $\pm$ 0.01 \\
\midrule
GCN & Mod + DGI\cite{velickovic2019deep} & 0.1 
& 0.752 $\pm$ 0.00 & 0.808 $\pm$ 0.01 & 0.728 $\pm$ 0.00 & 0.766 $\pm$ 0.00 \\
GAT\ & Mod + DGI & 0.9 
& 0.756 $\pm$ 0.01 & 0.809 $\pm$ 0.01 & 0.734 $\pm$ 0.01 & 0.770 $\pm$ 0.01 \\
ARMA & Mod + DGI & 0.5 
& 0.757 $\pm$ 0.01 & 0.813 $\pm$ 0.01 & 0.731 $\pm$ 0.00 & 0.770 $\pm$ 0.01 \\
\midrule
GCN & Mod + MERIT\cite{jin2021multi} & 0.1 
& 0.757 $\pm$ 0.00 & 0.814 $\pm$ 0.00 & 0.730 $\pm$ 0.00 & 0.770 $\pm$ 0.00 \\
GAT & Mod + MERIT & 0.9 
& 0.769 $\pm$ 0.01 & 0.826 $\pm$ 0.02 & 0.748 $\pm$ 0.01 & 0.784 $\pm$ 0.01 \\
\textbf{ARMA-C$^3$} & -- & 0.3
& \textbf{0.774 $\pm$ 0.01} & 0.835 $\pm$ 0.01 & 0.740 $\pm$ 0.00 & 0.785 $\pm$ 0.01 \\
Embedding & -- & - 
& \textbf{0.777 $\pm$ 0.00} & 0.843 $\pm$ 0.00 & 0.737 $\pm$ 0.00 & 0.786 $\pm$ 0.00 \\
\midrule
GCN & Cut\cite{bianchi2020spectral} & N/A 
& 0.761 $\pm$ 0.02 & 0.821 $\pm$ 0.02 & 0.731 $\pm$ 0.01 & 0.773 $\pm$ 0.01 \\
GAT & Cut & N/A 
& 0.755 $\pm$ 0.03 & 0.812 $\pm$ 0.03 & 0.728 $\pm$ 0.02 & 0.768 $\pm$ 0.03 \\
ARMA & Cut & N/A 
& 0.724 $\pm$ 0.09 & 0.783 $\pm$ 0.09 & 0.699 $\pm$ 0.08 & 0.738 $\pm$ 0.08 \\
\midrule
GCN & Cut + DGI & 0.3 
& 0.755 $\pm$ 0.02 & 0.813 $\pm$ 0.02 & 0.727 $\pm$ 0.02 & 0.768 $\pm$ 0.02 \\
GAT& Cut + DGI & 0.9 
& 0.751 $\pm$ 0.05 & 0.810 $\pm$ 0.05 & 0.723 $\pm$ 0.04 & 0.764 $\pm$ 0.05 \\
ARMA & Cut + DGI & 0.9 
& 0.744 $\pm$ 0.05 & 0.803 $\pm$ 0.05 & 0.717 $\pm$ 0.04 & 0.758 $\pm$ 0.05 \\
\midrule
GCN & Cut + MERIT & 0.005 
& 0.773 $\pm$ 0.01 & 0.832 $\pm$ 0.01 & 0.748 $\pm$ 0.00 & \textbf{0.786 $\pm$ 0.00} \\
GAT & Cut + MERIT & 0.5 
& 0.769 $\pm$ 0.01 & 0.827 $\pm$ 0.02 & 0.747 $\pm$ 0.01 & 0.783 $\pm$ 0.01 \\
ARMA & Cut + MERIT & 0.3 
& 0.769 $\pm$ 0.01 & 0.829 $\pm$ 0.01 & 0.744 $\pm$ 0.01 & 0.783 $\pm$ 0.01 \\
\bottomrule
\end{tabular}
}
\end{adjustbox}
\end{subtable}

\bigskip

\begin{subtable}{0.78\textwidth}
\centering
\caption{CN vs AD}
\setlength\tabcolsep{5pt}
\begin{adjustbox}{max width=\textwidth}
{\large
\begin{tabular}{@{} l l c c c c c @{}}
\toprule
\textbf{Model} & \textbf{Loss Function} & $\boldsymbol{\lambda_{con}}$ 
& \textbf{Accuracy} & \textbf{Precision} & \textbf{Recall} & \textbf{F1 Score} \\
\midrule
K-Means ~\cite{macqueen1967kmeans} & -- & -- 
& 0.771 $\pm$ 0.00 & 0.753 $\pm$ 0.01 & 0.644 $\pm$ 0.01 & 0.695 $\pm$ 0.00 \\
PCA + K-Means & -- & -- 
& 0.767 $\pm$ 0.00 
& 0.750 $\pm$ 0.00 
& 0.633 $\pm$ 0.00 
& 0.687 $\pm$ 0.00 \\
Token-Cut \cite{wang2023tokencut} & -- & -- 
& 0.753 $\pm$ 0.00 & 0.677 $\pm$ 0.00 & 0.744 $\pm$ 0.00 & 0.709 $\pm$ 0.00 \\
SpectralNet \cite{shaham2018spectralnet} & -- & -- 
& 0.769 $\pm$ 0.02 & 0.483 $\pm$ 0.28 & 0.483 $\pm$ 0.17 & 0.475 $\pm$ 0.23 \\
TANGO \cite{ma2024tango} & -- & -- 
& 0.516 $\pm$ 0.00 & 0.296 $\pm$ 0.00 & 0.144 $\pm$ 0.00 & 0.194 $\pm$ 0.00 \\
MAGI \cite{liu2024revisiting} & -- & -- 
& 0.665 $\pm$ 0.01 & \textbf{0.862 $\pm$ 0.03} & 0.203 $\pm$ 0.02 & 0.340 $\pm$ 0.03 \\

\midrule
GCN~\cite{kipf2017semi} & Mod~\cite{tsitsulin2023graph} & N/A 
& 0.765 $\pm$ 0.00 & 0.684 $\pm$ 0.00 & 0.777 $\pm$ 0.01 & 0.727 $\pm$ 0.00 \\

\textbf{GAT~\cite{velickovic2018graph}} & \textbf{Mod} & \textbf{N/A} 
& \textbf{0.785 $\pm$ 0.01} & 0.722 $\pm$ 0.01 & 0.762 $\pm$ 0.01 & 0.741 $\pm$ 0.00 \\

ARMA~\cite{bianchi2021graph} & Mod & N/A 
& 0.756 $\pm$ 0.01 & 0.669 $\pm$ 0.01 & 0.783 $\pm$ 0.01 & 0.722 $\pm$ 0.01 \\
\midrule
GCN & Mod + DGI\cite{velickovic2019deep} & 0.01 
& 0.773 $\pm$ 0.00 & 0.711 $\pm$ 0.00 & 0.737 $\pm$ 0.01 & 0.724 $\pm$ 0.01 \\
GAT & Mod + DGI & 1 
& 0.774 $\pm$ 0.01 & 0.726 $\pm$ 0.01 & 0.706 $\pm$ 0.03 & 0.715 $\pm$ 0.02 \\
ARMA & Mod + DGI & 0.09
& 0.779 $\pm$ 0.00 & 0.695 $\pm$ 0.00 & 0.803 $\pm$ 0.01 & \textbf{0.745 $\pm$ 0.00} \\
\midrule
GCN & Mod + MERIT\cite{jin2021multi} & 8
& 0.736 $\pm$ 0.01 & 0.632 $\pm$ 0.01 & \textbf{0.830 $\pm$ 0.01} & 0.717 $\pm$ 0.01 \\
GAT & Mod + MERIT & 0.5 
& 0.775 $\pm$ 0.01 & 0.712 $\pm$ 0.03 & 0.751 $\pm$ 0.04 & 0.729 $\pm$ 0.01 \\
ARMA-C$^3$ & -- & 0.5 
& 0.748 $\pm$ 0.01 & 0.642 $\pm$ 0.01 & 0.805 $\pm$ 0.01 & 0.712 $\pm$ 0.00 \\
Embedding & -- & - 
& 0.753 $\pm$ 0.00 & 0.658 $\pm$ 0.00 & 0.811 $\pm$ 0.00 & 0.726 $\pm$ 0.00 \\
\midrule
GCN & Cut\cite{bianchi2020spectral} & N/A 
& 0.769 $\pm$ 0.00 & 0.677 $\pm$ 0.00 & 0.817 $\pm$ 0.01 & 0.740 $\pm$ 0.01 \\
GAT & Cut & N/A 
& 0.770 $\pm$ 0.01 & 0.695 $\pm$ 0.01 & 0.766 $\pm$ 0.00 & 0.729 $\pm$ 0.01 \\
ARMA & Cut & N/A 
& 0.751 $\pm$ 0.01 & 0.655 $\pm$ 0.01 & 0.808 $\pm$ 0.01 & 0.723 $\pm$ 0.01 \\
\midrule
GCN & Cut + DGI & 0.01 
& 0.765 $\pm$ 0.00 & 0.687 $\pm$ 0.00 & 0.764 $\pm$ 0.01 & 0.724 $\pm$ 0.00 \\
GAT & Cut + DGI & 1 
& 0.774 $\pm$ 0.01 & 0.728 $\pm$ 0.01 & 0.704 $\pm$ 0.02 & 0.716 $\pm$ 0.01 \\
ARMA & Cut + DGI & 0.09 
& 0.754 $\pm$ 0.01 & 0.661 $\pm$ 0.01 & 0.801 $\pm$ 0.01 & 0.724 $\pm$ 0.01 \\
\midrule
GCN & Cut + MERIT & 8 
& 0.744 $\pm$ 0.00 & 0.649 $\pm$ 0.00 & 0.793 $\pm$ 0.01 & 0.714 $\pm$ 0.00 \\
GAT& Cut + MERIT & 1 
& 0.776 $\pm$ 0.01 & 0.718 $\pm$ 0.02 & 0.737 $\pm$ 0.05 & 0.725 $\pm$ 0.02 \\
ARMA & Cut + MERIT & 0.1 
& 0.746 $\pm$ 0.00 & 0.645 $\pm$ 0.01 & 0.799 $\pm$ 0.01 & 0.711 $\pm$ 0.01 \\
\bottomrule
\end{tabular}
}
\end{adjustbox}
\end{subtable}
\end{table*}
\begin{table*}[!htbp]
\centering
\caption{Comparison of unsupervised clustering performance on MedMNIST datasets using traditional and graph-based approaches.
Metrics are reported as mean $\pm$ standard deviation over ten repeated experiments.}
\label{tab:combined_medmnist_unsup}

\begin{subtable}{0.8\textwidth}
\centering
\caption{Clustering on BreastMNIST}
\setlength\tabcolsep{5pt}
\begin{adjustbox}{max width=\textwidth}
{\large
\begin{tabular}{@{} l l c c c c c @{}}
\toprule
\textbf{Model} & \textbf{Loss Function} & $\boldsymbol{\lambda_{con}}$ 
& \textbf{Accuracy} & \textbf{Precision} & \textbf{Recall} & \textbf{F1 Score} \\
\midrule
K-Means ~\cite{macqueen1967kmeans} & -- & -- 
& 0.541 $\pm$ 0.02 & 0.731 $\pm$ 0.07 & 0.619 $\pm$ 0.11 & 0.657 $\pm$ 0.06 \\
PCA + K-Means & -- & -- 
& 0.767 $\pm$ 0.00 
& 0.750 $\pm$ 0.00 
& 0.633 $\pm$ 0.00 
& 0.687 $\pm$ 0.00 \\
Token-Cut \cite{wang2023tokencut} & -- & -- 
& 0.580 $\pm$ 0.00 & \textbf{0.825 $\pm$ 0.00} & 0.539 $\pm$ 0.00 & 0.652 $\pm$ 0.00 \\
SpectralNet \cite{shaham2018spectralnet} & -- & -- 
& 0.616 $\pm$ 0.06 & 0.722 $\pm$ 0.09 & 0.472 $\pm$ 0.26 & 0.528 $\pm$ 0.22 \\
TANGO \cite{ma2024tango} & -- & -- 
& 0.580 $\pm$ 0.00 & \textbf{0.825 $\pm$ 0.00} & 0.539  $\pm$ 0.00 & 0.652 $\pm$ 0.00 \\
MAGI \cite{liu2024revisiting} & -- & -- 
& 0.661 $\pm$ 0.00 & 0.724 $\pm$ 0.00 & 0.867 $\pm$ 0.01 & 0.789 $\pm$ 0.00 \\
\midrule
GCN\cite{kipf2017semi} & Mod\cite{tsitsulin2023graph} & N/A 
& 0.576 $\pm$ 0.00 & 0.787 $\pm$ 0.01 & 0.576 $\pm$ 0.02 & 0.665 $\pm$ 0.01 \\
GAT\cite{velickovic2018graph} & Mod & N/A 
& 0.575 $\pm$ 0.02 & 0.806 $\pm$ 0.01 & 0.552 $\pm$ 0.03 & 0.655 $\pm$ 0.02 \\
ARMA\cite{bianchi2021graph} & Mod & N/A 
& 0.580 $\pm$ 0.04 & 0.780 $\pm$ 0.03 & 0.601 $\pm$ 0.11 & 0.670 $\pm$ 0.07 \\
\midrule
GCN  & Mod + DGI\cite{velickovic2019deep} & 0.001 
& 0.601 $\pm$ 0.02 & 0.633 $\pm$ 0.01 & 0.421 $\pm$ 0.03 & 0.506 $\pm$ 0.02 \\
GAT & Mod + DGI & 0.001 
& 0.650 $\pm$ 0.05 & 0.596 $\pm$ 0.30 & 0.493 $\pm$ 0.35 & 0.501 $\pm$ 0.31 \\
ARMA & Mod + DGI & 0.01 
& 0.712 $\pm$ 0.01 & 0.678 $\pm$ 0.10 & 0.581 $\pm$ 0.44 & 0.523 $\pm$ 0.37 \\
\midrule
GCN & Mod + MERIT\cite{jin2021multi} & 0.01 
& 0.619 $\pm$ 0.05 & 0.724 $\pm$ 0.01 & 0.771 $\pm$ 0.09 & 0.745 $\pm$ 0.05 \\
GAT & Mod + MERIT & 0.5 
& 0.610 $\pm$ 0.05 & 0.788 $\pm$ 0.05 & 0.647 $\pm$ 0.10 & 0.705 $\pm$ 0.05 \\
\textbf{ARMA-C$^3$} & \textbf{--} & \textbf{0.001} 
& \textbf{0.731 $\pm$ 0.00} & 0.731 $\pm$ 0.00 & \textbf{0.979 $\pm$ 0.01} & \textbf{0.844 $\pm$ 0.00} \\
Embedding & -- & - 
& 0.728 $\pm$ 0.00 & 0.727 $\pm$ 0.00 & 0.952 $\pm$ 0.00 & 0.839 $\pm$ 0.00 \\
\midrule
GCN & Cut\cite{bianchi2020spectral} & N/A 
& 0.545 $\pm$ 0.03 & 0.776 $\pm$ 0.02 & 0.530 $\pm$ 0.02 & 0.630 $\pm$ 0.02 \\
GAT & Cut & N/A 
& 0.549 $\pm$ 0.03 & 0.781 $\pm$ 0.03 & 0.533 $\pm$ 0.02 & 0.633 $\pm$ 0.02 \\
ARMA & Cut & N/A 
& 0.580 $\pm$ 0.01 & 0.760 $\pm$ 0.01 & 0.622 $\pm$ 0.03 & 0.684 $\pm$ 0.01 \\
\midrule
GCN & Cut + DGI & 0.001 
& 0.604 $\pm$ 0.14 & 0.762 $\pm$ 0.05 & 0.674 $\pm$ 0.28 & 0.670 $\pm$ 0.22 \\
GAT & Cut + DGI & 0.001 
& 0.620 $\pm$ 0.07 & 0.681 $\pm$ 0.13 & 0.356 $\pm$ 0.25 & 0.418 $\pm$ 0.26 \\
ARMA & Cut + DGI & 0.01 
& 0.660 $\pm$ 0.03 & 0.733 $\pm$ 0.01 & 0.547 $\pm$ 0.36 & 0.544 $\pm$ 0.29 \\
\midrule
GCN & Cut + MERIT & 0.01 
& 0.620 $\pm$ 0.04 & 0.731 $\pm$ 0.04 & 0.768 $\pm$ 0.09 & 0.745 $\pm$ 0.04 \\
GAT & Cut + MERIT & 0.5 
& 0.628 $\pm$ 0.07 & 0.788 $\pm$ 0.04 & 0.678 $\pm$ 0.13 & 0.720 $\pm$ 0.07 \\
ARMA & Cut + MERIT & 0.001
& 0.722 $\pm$ 0.01 & 0.733 $\pm$ 0.00 & 0.977 $\pm$ 0.01 & 0.837 $\pm$ 0.01 \\
\bottomrule
\end{tabular}
}
\end{adjustbox}
\end{subtable}

\bigskip

\begin{subtable}{0.8\textwidth}
\centering
\caption{Clustering on PneumoniaMNIST}
\setlength\tabcolsep{5pt}
\begin{adjustbox}{max width=\textwidth}
{\large
\begin{tabular}{@{} l l c c c c c @{}}
\toprule
\textbf{Model} & \textbf{Loss Function} & $\boldsymbol{\lambda_{con}}$ 
& \textbf{Accuracy} & \textbf{Precision} & \textbf{Recall} & \textbf{F1 Score} \\
\midrule
K-Means ~\cite{macqueen1967kmeans} & -- & -- 
& 0.899 $\pm$ 0.00 & 0.969 $\pm$ 0.00 & 0.847 $\pm$ 0.00 & 0.904 $\pm$ 0.00 \\
PCA + K-Means & -- & -- 
& 0.888 $\pm$ 0.00 
& 0.961 $\pm$ 0.01 
& 0.832 $\pm$ 0.01 
& 0.892 $\pm$ 0.00 \\
Token-Cut \cite{wang2023tokencut} & -- & -- 
& 0.885 $\pm$ 0.00 & 0.954 $\pm$ 0.00 & 0.834 $\pm$ 0.00 & 0.890 $\pm$ 0.00 \\
SpectralNet \cite{shaham2018spectralnet} & -- & -- 
& 0.903 $\pm$ 0.01 & 0.569 $\pm$ 0.41 & 0.492 $\pm$ 0.35 & 0.527 $\pm$ 0.38 \\
TANGO \cite{ma2024tango} & -- & -- 
& 0.900 $\pm$ 0.00 & 0.961 $\pm$ 0.00 & 0.855 $\pm$ 0.00 & 0.905 $\pm$ 0.00 \\
\textbf{MAGI \cite{liu2024revisiting}} & -- & -- 
& 0.894 $\pm$ 0.00 & 0.897 $\pm$ 0.00 & \textbf{0.916 $\pm$ 0.00} & 0.906 $\pm$ 0.00 \\

\midrule
GCN~\cite{kipf2017semi} & Mod~\cite{tsitsulin2023graph} & N/A 
& 0.914 $\pm$ 0.00 & 0.962 $\pm$ 0.00 & 0.880 $\pm$ 0.00 & 0.919 $\pm$ 0.00 \\

GAT~\cite{velickovic2018graph} & Mod & N/A 
& 0.912 $\pm$ 0.00 & 0.961 $\pm$ 0.00 & 0.877 $\pm$ 0.00 & 0.917 $\pm$ 0.00 \\

ARMA~\cite{bianchi2021graph} & Mod & N/A 
& 0.909 $\pm$ 0.00 & 0.961 $\pm$ 0.00 & 0.872 $\pm$ 0.00 & 0.914 $\pm$ 0.00 \\
\midrule
GCN & Mod + DGI\cite{velickovic2019deep} & 0.001 
& 0.914 $\pm$ 0.00 & 0.959 $\pm$ 0.00 & 0.884 $\pm$ 0.01 & 0.920 $\pm$ 0.00 \\
GAT & Mod + DGI & 0.05 
& 0.911 $\pm$ 0.01 & 0.956 $\pm$ 0.01 & 0.882 $\pm$ 0.02 & 0.917 $\pm$ 0.01 \\
ARMA & Mod + DGI & 0.3 
& 0.914 $\pm$ 0.00 & 0.963 $\pm$ 0.00 & 0.879 $\pm$ 0.00 & 0.919 $\pm$ 0.00 \\
\midrule
\textbf{GCN} & \textbf{Mod + MERIT\cite{jin2021multi}} & \textbf{0.005} 
& \textbf{0.919 $\pm$ 0.00} & 0.945 $\pm$ 0.00 & 0.908 $\pm$ 0.01 & \textbf{0.926 $\pm$ 0.00} \\
GAT & Mod + MERIT & 0.009 
& 0.914 $\pm$ 0.00 & 0.947 $\pm$ 0.01 & 0.895 $\pm$ 0.01 & 0.921 $\pm$ 0.00 \\
ARMA-C$^3$ & -- & 5 
& 0.910 $\pm$ 0.00 & \textbf{0.971 $\pm$ 0.00} & 0.865 $\pm$ 0.00 & 0.915 $\pm$ 0.00 \\
Embedding & -- & - 
& 0.914 $\pm$ 0.00 & 0.965 $\pm$ 0.00 & 0.860 $\pm$ 0.00 & 0.909 $\pm$ 0.00 \\
\midrule
GCN  & Cut\cite{bianchi2020spectral} & N/A 
& 0.913 $\pm$ 0.01 & 0.971 $\pm$ 0.00 & 0.871 $\pm$ 0.00 & 0.918 $\pm$ 0.00 \\
GAT & Cut & N/A 
& 0.906 $\pm$ 0.00 & 0.964 $\pm$ 0.00 & 0.864 $\pm$ 0.00 & 0.911 $\pm$ 0.00 \\
ARMA & Cut & N/A 
& 0.904 $\pm$ 0.00 & 0.962 $\pm$ 0.00 & 0.862 $\pm$ 0.00 & 0.909 $\pm$ 0.00 \\
\midrule
GCN & Cut + DGI & 0.1 
& 0.901 $\pm$ 0.00 & 0.961 $\pm$ 0.00 & 0.858 $\pm$ 0.00 & 0.907 $\pm$ 0.00 \\
GAT & Cut + DGI & 0.1 
& 0.911 $\pm$ 0.00 & 0.954 $\pm$ 0.01 & 0.883 $\pm$ 0.01 & 0.917 $\pm$ 0.00 \\
ARMA & Cut + DGI & 0.1 
& 0.915 $\pm$ 0.00 & 0.970 $\pm$ 0.00 & 0.875 $\pm$ 0.00 & 0.920 $\pm$ 0.00 \\
\midrule
GCN & Cut + MERIT & 0.0001 
& 0.809 $\pm$ 0.03 & 0.986 $\pm$ 0.00 & 0.667 $\pm$ 0.05 & 0.795 $\pm$ 0.04 \\
GAT & Cut + MERIT & 0.005 
& 0.900 $\pm$ 0.00 & 0.905 $\pm$ 0.00 & 0.905 $\pm$ 0.00 & 0.905 $\pm$ 0.00 \\
ARMA & Cut + MERIT & 0.005 
& 0.905 $\pm$ 0.00 & 0.965 $\pm$ 0.00 & 0.861 $\pm$ 0.00 & 0.910 $\pm$ 0.00 \\
\bottomrule
\end{tabular}
}
\end{adjustbox}
\end{subtable}

\end{table*}
\begin{table*}[!htbp]
\centering
\caption{Comparison of (a) unsupervised clustering and (b) semi-supervised classification performance on the NIFD dataset.
Unsupervised results are reported as mean $\pm$ standard deviation over ten repeated runs,
while semi-supervised results are averaged over 20-fold cross-validation.}

\label{tab:nifd_combined}

\begin{subtable}{0.83\textwidth}
\centering
\caption{Clustering on NIFD}
\setlength\tabcolsep{5pt}
\begin{adjustbox}{max width=\textwidth}
{\large
\begin{tabular}{@{} l l c c c c c @{}}
\toprule
\textbf{Model} & \textbf{Loss Function} & $\boldsymbol{\lambda_{con}}$ 
& \textbf{Accuracy} & \textbf{Precision} & \textbf{Recall} & \textbf{F1 Score} \\
\midrule
\textbf{K-Means ~\cite{macqueen1967kmeans}} & \textbf{--} & -- 
& 0.519 $\pm$ 0.00 & \textbf{0.966 $\pm$ 0.00} & 0.294 $\pm$ 0.00 & 0.451 $\pm$ 0.00 \\
PCA + K-Means & -- & -- 
& 0.501 $\pm$ 0.00 
& 0.967 $\pm$ 0.00 
& 0.296 $\pm$ 0.00 
& 0.453 $\pm$ 0.00 \\
Token-Cut \cite{wang2023tokencut} & -- & -- 
& 0.569 $\pm$ 0.00 & 0.857 $\pm$ 0.00 & 0.429 $\pm$ 0.00 & 0.571 $\pm$ 0.00 \\
SpectralNet \cite{shaham2018spectralnet} & -- & -- 
& 0.533 $\pm$ 0.01 & 0.862 $\pm$ 0.16 & 0.404 $\pm$ 0.14 & 0.516 $\pm$ 0.06 \\
TANGO \cite{ma2024tango} & -- & -- 
& 0.548 $\pm$ 0.00 & 0.420 $\pm$ 0.00 & 0.979 $\pm$ 0.00 & 0.588 $\pm$ 0.00 \\
MAGI\cite{liu2024revisiting} & -- & -- 
& 0.665 $\pm$ 0.01 & 0.862 $\pm$ 0.03 & 0.203 $\pm$ 0.02 & 0.328 $\pm$ 0.03 \\

\midrule
GCN~\cite{kipf2017semi} & Mod~\cite{tsitsulin2023graph} & N/A 
& 0.684 $\pm$ 0.00 & 0.706 $\pm$ 0.00 & 0.907 $\pm$ 0.00 & 0.794 $\pm$ 0.00 \\

GAT~\cite{velickovic2018graph} & Mod & N/A 
& 0.692 $\pm$ 0.02 & 0.807 $\pm$ 0.05 & 0.724 $\pm$ 0.08 & 0.758 $\pm$ 0.02 \\

ARMA~\cite{bianchi2021graph} & Mod & N/A 
& 0.692 $\pm$ 0.02 & 0.862 $\pm$ 0.02 & 0.645 $\pm$ 0.01 & 0.738 $\pm$ 0.01 \\
\midrule
GCN & Mod + DGI\cite{velickovic2019deep} & 0.1 
& 0.683 $\pm$ 0.00 & 0.704 $\pm$ 0.00 & 0.911 $\pm$ 0.00 & 0.794 $\pm$ 0.00 \\
GAT & Mod + DGI & 2 
& 0.677 $\pm$ 0.02 & 0.718 $\pm$ 0.04 & 0.876 $\pm$ 0.11 & 0.781 $\pm$ 0.04 \\
ARMA & Mod + DGI & 8 
& 0.710 $\pm$ 0.02 & 0.882 $\pm$ 0.02 & 0.655 $\pm$ 0.02 & 0.752 $\pm$ 0.01 \\
\midrule
GCN & Mod + MERIT\cite{jin2021multi} & 0.5 
& 0.672 $\pm$ 0.01 & 0.673 $\pm$ 0.01 & 0.996 $\pm$ 0.01 & \textbf{0.803 $\pm$ 0.01} \\
GAT & Mod + MERIT & 0.05 
& 0.671 $\pm$ 0.00 & 0.671 $\pm$ 0.00 & \textbf{1.000 $\pm$ 0.00} & \textbf{0.803 $\pm$ 0.00} \\
\textbf{ARMA-C$^3$} & \textbf{--} & 0.01 
& \textbf{0.712 $\pm$ 0.01} & 0.887 $\pm$ 0.01 & 0.638 $\pm$ 0.01 & 0.742 $\pm$ 0.01 \\
Embedding & -- & - 
& 0.719 $\pm$ 0.00 & 0.886 $\pm$ 0.00 & 0.633 $\pm$ 0.00 & 0.738 $\pm$ 0.00 \\
\midrule
GCN & Cut\cite{bianchi2020spectral} & N/A 
& 0.672 $\pm$ 0.01 & 0.694 $\pm$ 0.02 & 0.917 $\pm$ 0.02 & 0.790 $\pm$ 0.00 \\
GAT & Cut & N/A 
& 0.671 $\pm$ 0.02 & 0.694 $\pm$ 0.01 & 0.916 $\pm$ 0.06 & 0.788 $\pm$ 0.02 \\
ARMA & Cut & N/A 
& 0.608 $\pm$ 0.05 & 0.779 $\pm$ 0.05 & 0.581 $\pm$ 0.04 & 0.665 $\pm$ 0.05 \\
\midrule
GCN & Cut + DGI & 0.1 
& 0.679 $\pm$ 0.02 & 0.683 $\pm$ 0.01 & 0.975 $\pm$ 0.04 & \textbf{0.803 $\pm$ 0.01} \\
GAT & Cut + DGI & 2 
& 0.668 $\pm$ 0.01 & 0.683 $\pm$ 0.02 & 0.946 $\pm$ 0.04 & 0.793 $\pm$ 0.01 \\
ARMA & Cut + DGI & 1 
& 0.670 $\pm$ 0.02 & 0.844 $\pm$ 0.02 & 0.624 $\pm$ 0.02 & 0.717 $\pm$ 0.02 \\
\midrule
GCN & Cut + MERIT & 5 
& 0.670 $\pm$ 0.01 & 0.676 $\pm$ 0.01 & 0.978 $\pm$ 0.03 & 0.799 $\pm$ 0.01 \\
GAT & Cut + MERIT & 0.01 
& 0.640 $\pm$ 0.07 & 0.719 $\pm$ 0.08 & 0.808 $\pm$ 0.18 & 0.742 $\pm$ 0.07 \\
ARMA & Cut + MERIT & 0.01 
& 0.564 $\pm$ 0.06 & 0.740 $\pm$ 0.05 & 0.540 $\pm$ 0.05 & 0.624 $\pm$ 0.05 \\
\bottomrule
\end{tabular}
}
\end{adjustbox}
\end{subtable}

\bigskip

\begin{subtable}{0.88\textwidth}
\centering
\caption{semi-supervised classification on NIFD (10\% Training, 90\% Testing)}
\setlength\tabcolsep{5pt}
\begin{adjustbox}{max width=\textwidth}
{\large
\begin{tabular}{@{} l l c c c c c @{}}
\toprule
\textbf{Model} & \textbf{Loss Function} & $\boldsymbol{\lambda_{con}}$ 
& \textbf{Accuracy} & \textbf{Precision} & \textbf{Recall} & \textbf{F1 Score} \\
\midrule
SVC\cite{cortes1995support} & -- & -- 
& 0.708 $\pm$ 0.03 & 0.730 $\pm$ 0.03 & \textbf{0.903 $\pm$ 0.05} & \textbf{0.806 $\pm$ 0.02} \\

XGBoost\cite{chen2016xgboost} & -- & -- 
& 0.604 $\pm$ 0.10 & 0.703 $\pm$ 0.07 & 0.711 $\pm$ 0.19 & 0.694 $\pm$ 0.12 \\

MLPClassifier\cite{rumelhart1986learning} & -- & -- 
& 0.681 $\pm$ 0.05 & 0.790 $\pm$ 0.05 & 0.723 $\pm$ 0.06 & 0.753 $\pm$ 0.04 \\

Random Forest\cite{breiman2001random} & -- & -- 
& 0.702 $\pm$ 0.03 & 0.729 $\pm$ 0.03 & 0.894 $\pm$ 0.08 & 0.800 $\pm$ 0.03 \\

APPNP\cite{gasteiger2018predict} & -- & -- 
& 0.690 $\pm$ 0.04 & 0.781 $\pm$ 0.04 & 0.757 $\pm$ 0.08 & 0.766 $\pm$ 0.04 \\

JacobiConv\cite{wang2022powerful} & -- & -- 
& 0.692 $\pm$ 0.05 & 0.778 $\pm$ 0.05 & 0.769 $\pm$ 0.09 & 0.769 $\pm$ 0.05 \\
\midrule
ARMA~\cite{bianchi2021graph} & Baseline & -- 
& 0.630 $\pm$ 0.05 & 0.785 $\pm$ 0.06 & 0.625 $\pm$ 0.04 & 0.695 $\pm$ 0.04 \\

GAT\cite{velickovic2018graph} & Baseline & -- 
& 0.639 $\pm$ 0.06 & 0.806 $\pm$ 0.08 & 0.621 $\pm$ 0.04 & 0.699 $\pm$ 0.04 \\
\midrule
ARMA & Mod~\cite{tsitsulin2023graph} & 0.01 
& 0.713 $\pm$ 0.04 & 0.764 $\pm$ 0.05 & 0.846 $\pm$ 0.10 & 0.796 $\pm$ 0.05 \\

GAT & Mod & 0.5 
& 0.700 $\pm$ 0.05 & 0.747 $\pm$ 0.06 & 0.858 $\pm$ 0.11 & 0.791 $\pm$ 0.05 \\

ARMA & Cut\cite{bianchi2020spectral} & 0.9 
& 0.675 $\pm$ 0.05 & \textbf{0.847 $\pm$ 0.05} & 0.634 $\pm$ 0.04 & 0.725 $\pm$ 0.04 \\

GAT & Cut & 0.5 
& 0.682 $\pm$ 0.05 & 0.856 $\pm$ 0.05 & 0.635 $\pm$ 0.04 & 0.729 $\pm$ 0.05 \\
\midrule
ARMA & DGI\cite{velickovic2019deep} + Mod & 0.009 
& 0.681 $\pm$ 0.06 & 0.776 $\pm$ 0.07 & 0.751 $\pm$ 0.12 & 0.755 $\pm$ 0.06 \\

GAT & DGI + Mod & 0.001 
& 0.656 $\pm$ 0.06 & 0.793 $\pm$ 0.06 & 0.659 $\pm$ 0.07 & 0.718 $\pm$ 0.06 \\

ARMA & DGI + Cut & 0.001 
& 0.673 $\pm$ 0.06 & 0.770 $\pm$ 0.07 & 0.755 $\pm$ 0.18 & 0.744 $\pm$ 0.09 \\

GAT & DGI + Cut & 0.001 
& 0.637 $\pm$ 0.06 & 0.819 $\pm$ 0.07 & 0.587 $\pm$ 0.05 & 0.684 $\pm$ 0.06 \\
\midrule
\textbf{ARMA-C$^3$} & \textbf{--} & \textbf{0.01} 
& \textbf{0.723 $\pm$ 0.05} & 0.812 $\pm$ 0.05 & 0.771 $\pm$ 0.07 & 0.789 $\pm$ 0.04 \\

GAT & MERIT\cite{jin2021multi} + Mod & 0.001 
& 0.711 $\pm$ 0.05 & 0.806 $\pm$ 0.05 & 0.757 $\pm$ 0.07 & 0.778 $\pm$ 0.04 \\

\textbf{ARMA} & \textbf{MERIT + Cut} & \textbf{0.01} 
& \textbf{0.723 $\pm$ 0.05} & 0.812 $\pm$ 0.05 & 0.773 $\pm$ 0.07 & 0.789 $\pm$ 0.04 \\

GAT & MERIT + Cut & 0.0001 
& 0.703 $\pm$ 0.05 & 0.800 $\pm$ 0.05 & 0.753 $\pm$ 0.06 & 0.773 $\pm$ 0.04 \\
\bottomrule
\end{tabular}
}
\end{adjustbox}
\end{subtable}
\end{table*}

\begin{table*}[!htbp]
\centering
\caption{Comparison of (a) unsupervised clustering and (b)semi-supervised classification performance on the liver ultrasound dataset.
Unsupervised results are reported as mean $\pm$ standard deviation over ten repeated runs, while semi-supervised results are averaged over 20-fold cross-validation.}
\label{tab:liver_unsup_results}

\begin{subtable}{0.8\textwidth}
\centering
\caption{Clustering on liver ultrasound dataset}
\setlength\tabcolsep{5pt}
\begin{adjustbox}{max width=\textwidth}
{\large
\begin{tabular}{@{} l l c c c c c @{}}
\toprule
\textbf{Model} & \textbf{Loss Function} & $\boldsymbol{\lambda_{con}}$
& \textbf{Accuracy} & \textbf{Precision} & \textbf{Recall} & \textbf{F1 Score} \\
\midrule

K-Means~\cite{macqueen1967kmeans} & -- & -- 
& 0.712 $\pm$ 0.00 & 0.817 $\pm$ 0.00 & 0.748 $\pm$ 0.00 & 0.781 $\pm$ 0.00 \\
PCA + K-Means & -- & -- 
& 0.713 $\pm$ 0.00 
& 0.817 $\pm$ 0.00 
& 0.749 $\pm$ 0.00 
& 0.782 $\pm$ 0.00 \\
SpectralNet~\cite{shaham2018spectralnet} & -- & -- 
& 0.725 $\pm$ 0.05 & 0.551 $\pm$ 0.18 & 0.420 $\pm$ 0.28 & 0.462 $\pm$ 0.25 \\

TokenCut~\cite{wang2023tokencut} & -- & -- 
& 0.666 $\pm$ 0.00 & 0.829 $\pm$ 0.00 & 0.646 $\pm$ 0.00 & 0.726 $\pm$ 0.00 \\

TANGO~\cite{ma2024tango} & -- & -- 
& 0.694 $\pm$ 0.00 & 0.763 $\pm$ 0.00 & 0.710 $\pm$ 0.00 & 0.736 $\pm$ 0.00 \\

MAGI~\cite{liu2024revisiting} & -- & -- 
& 0.598 $\pm$ 0.06 & 0.822 $\pm$ 0.01 & 0.528 $\pm$ 0.12 & 0.634 $\pm$ 0.09 \\

\midrule

GCN~\cite{kipf2017semi} & Mod~\cite{tsitsulin2023graph} & N/A 
& 0.692 $\pm$ 0.00 & 0.852 $\pm$ 0.00 & 0.665 $\pm$ 0.00 & 0.747 $\pm$ 0.00 \\

GAT~\cite{velickovic2018graph} & Mod & N/A 
& 0.716 $\pm$ 0.00 & \textbf{0.870 $\pm$ 0.00} & 0.688 $\pm$ 0.00 & 0.768 $\pm$ 0.00 \\

ARMA~\cite{bianchi2021graph} & Mod & N/A 
& 0.721 $\pm$ 0.01 & 0.836 $\pm$ 0.00 & 0.738 $\pm$ 0.01 & 0.784 $\pm$ 0.01 \\

\midrule

GCN & Mod + DGI~\cite{velickovic2019deep} & 0.001 
& 0.736 $\pm$ 0.01 & 0.856 $\pm$ 0.01 & 0.738 $\pm$ 0.02 & 0.793 $\pm$ 0.01 \\

GAT & Mod + DGI & 8 
& 0.744 $\pm$ 0.02 & 0.857 $\pm$ 0.02 & 0.753 $\pm$ 0.05 & 0.800 $\pm$ 0.02 \\

ARMA & Mod + DGI & 0.9 
& 0.747 $\pm$ 0.00 & 0.830 $\pm$ 0.00 & 0.794 $\pm$ 0.01 & 0.811 $\pm$ 0.00 \\

\midrule

GCN & Mod + MERIT\cite{jin2021multi} & 0.05 
& 0.747 $\pm$ 0.00 & 0.825 $\pm$ 0.01 & 0.800 $\pm$ 0.01 & 0.812 $\pm$ 0.00 \\

GAT & Mod + MERIT & 2 
& 0.720 $\pm$ 0.04 & 0.857 $\pm$ 0.01 & 0.712 $\pm$ 0.08 & 0.775 $\pm$ 0.05 \\

\textbf{ARMA-C$^3$} & \textbf{--} & \textbf{8} 
& \textbf{0.754 $\pm$ 0.00} & 0.835 $\pm$ 0.00 & \textbf{0.800 $\pm$ 0.00} & \textbf{0.817 $\pm$ 0.00} \\

\midrule

GCN & Cut~\cite{bianchi2020spectral} & N/A 
& 0.640 $\pm$ 0.07 & 0.821 $\pm$ 0.06 & 0.605 $\pm$ 0.05 & 0.697 $\pm$ 0.06 \\

GAT & Cut & N/A 
& 0.654 $\pm$ 0.05 & 0.831 $\pm$ 0.04 & 0.622 $\pm$ 0.04 & 0.711 $\pm$ 0.04 \\

ARMA & Cut & N/A 
& 0.707 $\pm$ 0.00 & 0.810 $\pm$ 0.01 & 0.748 $\pm$ 0.02 & 0.778 $\pm$ 0.01 \\

\midrule

GCN & Cut + DGI & 0.001 
& 0.686 $\pm$ 0.07 & 0.831 $\pm$ 0.06 & 0.678 $\pm$ 0.07 & 0.746 $\pm$ 0.06 \\

GAT & Cut + DGI & 0.001 
& 0.707 $\pm$ 0.03 & 0.841 $\pm$ 0.01 & 0.707 $\pm$ 0.07 & 0.766 $\pm$ 0.04 \\

ARMA & Cut + DGI & 0.09 
& 0.726 $\pm$ 0.01 & 0.833 $\pm$ 0.01 & 0.752 $\pm$ 0.02 & 0.790 $\pm$ 0.01 \\

\midrule

GCN & Cut + MERIT & 0.05 
& 0.704 $\pm$ 0.03 & 0.855 $\pm$ 0.01 & 0.683 $\pm$ 0.04 & 0.759 $\pm$ 0.03 \\

GAT & Cut + MERIT & 0.05 
& 0.705 $\pm$ 0.03 & 0.791 $\pm$ 0.01 & 0.776 $\pm$ 0.06 & 0.782 $\pm$ 0.03 \\

ARMA & Cut + MERIT & 8 
& 0.728 $\pm$ 0.01 & 0.822 $\pm$ 0.03 & 0.774 $\pm$ 0.04 & 0.796 $\pm$ 0.01 \\

\bottomrule
\end{tabular}
}
\end{adjustbox}
\end{subtable}

\bigskip

\begin{subtable}{0.88\textwidth}
\centering
\caption{semi-supervised classification on the liver ultrasound dataset (10\% Training, 90\% Testing)}
\setlength\tabcolsep{5pt}
\begin{adjustbox}{max width=\textwidth}
{\large
\begin{tabular}{@{} l l c c c c c @{}}
\toprule
\textbf{Model} & \textbf{Loss Function} & $\boldsymbol{\lambda_{con}}$
& \textbf{Accuracy} & \textbf{Precision} & \textbf{Recall} & \textbf{F1 Score} \\
\midrule

SVC\cite{cortes1995support} & -- & -- 
& 0.742 $\pm$ 0.02 & 0.767 $\pm$ 0.02 & 0.897 $\pm$ 0.04 & 0.826 $\pm$ 0.01 \\

XGBoost\cite{chen2016xgboost} & -- & -- 
& 0.715 $\pm$ 0.02 & 0.760 $\pm$ 0.02 & 0.855 $\pm$ 0.04 & 0.804 $\pm$ 0.02 \\

MLPClassifier\cite{rumelhart1986learning} & -- & -- 
& 0.730 $\pm$ 0.03 & 0.817 $\pm$ 0.03 & 0.782 $\pm$ 0.04 & 0.799 $\pm$ 0.02 \\

Random Forest\cite{breiman2001random} & -- & -- 
& 0.735 $\pm$ 0.01 & 0.751 $\pm$ 0.02 & \textbf{0.921 $\pm$ 0.02} & 0.827 $\pm$ 0.00 \\

APPNP\cite{gasteiger2018predict} & -- & -- 
& 0.737 $\pm$ 0.02 & 0.784 $\pm$ 0.03 & 0.852 $\pm$ 0.04 & 0.816 $\pm$ 0.02 \\

JacobiConv\cite{wang2022powerful} & -- & -- 
& 0.726 $\pm$ 0.04 & 0.796 $\pm$ 0.03 & 0.813 $\pm$ 0.10 & 0.800 $\pm$ 0.05 \\

\midrule

ARMA\cite{bianchi2021graph} & Baseline & -- 
& 0.737 $\pm$ 0.03 & 0.792 $\pm$ 0.02 & 0.837 $\pm$ 0.06 & 0.813 $\pm$ 0.03 \\

GAT\cite{velickovic2019deep} & Baseline & -- 
& 0.704 $\pm$ 0.03 & 0.783 $\pm$ 0.02 & 0.788 $\pm$ 0.06 & 0.784 $\pm$ 0.03 \\

\midrule

ARMA & Mod\cite{tsitsulin2023graph} & -- 
& 0.742 $\pm$ 0.02 & 0.791 $\pm$ 0.02 & 0.851 $\pm$ 0.05 & 0.818 $\pm$ 0.02 \\

GAT & Mod & -- 
& 0.702 $\pm$ 0.03 & 0.781 $\pm$ 0.02 & 0.767 $\pm$ 0.06 & 0.773 $\pm$ 0.03 \\

ARMA & Cut\cite{bianchi2020spectral} & -- 
& 0.734 $\pm$ 0.03 & 0.789 $\pm$ 0.02 & 0.837 $\pm$ 0.05 & 0.811 $\pm$ 0.03 \\

GAT & Cut & -- 
& 0.710 $\pm$ 0.04 & 0.808 $\pm$ 0.04 & 0.761 $\pm$ 0.07 & 0.781 $\pm$ 0.04 \\

\midrule

ARMA & DGI~\cite{velickovic2019deep} + Mod & 0.01 
& 0.745 $\pm$ 0.02 & \textbf{0.824 $\pm$ 0.03} & 0.801 $\pm$ 0.05 & 0.811 $\pm$ 0.02 \\

GAT & DGI + Mod & 0.5 
& 0.712 $\pm$ 0.04 & \textbf{0.847 $\pm$ 0.04} & 0.693 $\pm$ 0.06 & 0.760 $\pm$ 0.04 \\

ARMA & DGI + Cut & 0.01 
& 0.729 $\pm$ 0.02 & 0.841 $\pm$ 0.03 & 0.749 $\pm$ 0.05 & 0.790 $\pm$ 0.02 \\

GAT & DGI + Cut & 0.001 
& 0.708 $\pm$ 0.05 & 0.820 $\pm$ 0.04 & 0.704 $\pm$ 0.10 & 0.752 $\pm$ 0.06 \\

\midrule

\textbf{ARMA-C$^3$} & \textbf{--} & \textbf{0.001} 
& \textbf{0.785 $\pm$ 0.02} & 0.818 $\pm$ 0.03 & 0.866 $\pm$ 0.04 & \textbf{0.841 $\pm$ 0.01} \\

GAT & MERIT\cite{jin2021multi} + Mod & 0.01 
& 0.734 $\pm$ 0.03 & 0.816 $\pm$ 0.03 & 0.794 $\pm$ 0.04 & 0.803 $\pm$ 0.02 \\

ARMA & MERIT + Cut & 0.001 
& 0.771 $\pm$ 0.01 & 0.806 $\pm$ 0.02 & 0.857 $\pm$ 0.03 & \textbf{0.840 $\pm$ 0.01} \\

GAT & MERIT + Cut & 0.01 
& 0.726 $\pm$ 0.05 & 0.807 $\pm$ 0.03 & 0.790 $\pm$ 0.08 & 0.796 $\pm$ 0.05 \\

\bottomrule
\end{tabular}
}
\end{adjustbox}
\end{subtable}
\end{table*}
\begin{table*}[!htbp]
\centering
\caption{Comparison of semi-supervised classification performance on the ADNI dataset. Metrics are reported as mean $\pm$ standard deviation over 20-fold cross-validation.}
\label{tab:adni_combined}

\begin{subtable}{0.97\textwidth}
\centering
\caption{semi-supervised classification (CN vs. MCI, 10\% Training, 90\% Testing)}
\setlength\tabcolsep{5pt}
\begin{adjustbox}{max width=\textwidth}
{\small
\begin{tabular}{@{} l l c c c c c @{}}
\toprule
\textbf{Model} & \textbf{Loss Function} & $\boldsymbol{\lambda_{con}}$
& \textbf{Accuracy} & \textbf{Precision} & \textbf{Recall} & \textbf{F1 Score} \\
\midrule

SVC\cite{cortes1995support} & -- & -- 
& 0.772 $\pm$ 0.02 & 0.792 $\pm$ 0.04 & 0.807 $\pm$ 0.05 & 0.797 $\pm$ 0.02 \\

XGBoost\cite{chen2016xgboost} & -- & -- 
& 0.739 $\pm$ 0.03 & 0.756 $\pm$ 0.04 & 0.790 $\pm$ 0.07 & 0.770 $\pm$ 0.03 \\

MLPClassifier\cite{rumelhart1986learning} & -- & -- 
& 0.783 $\pm$ 0.02 & 0.810 $\pm$ 0.03 & 0.802 $\pm$ 0.04 & 0.804 $\pm$ 0.02 \\

Random Forest\cite{breiman2001random} & -- & -- 
& 0.783 $\pm$ 0.02 & 0.800 $\pm$ 0.03 & 0.818 $\pm$ 0.06 & 0.806 $\pm$ 0.02 \\

APPNP\cite{gasteiger2018predict} & -- & -- 
& 0.782 $\pm$ 0.02 & 0.807 $\pm$ 0.05 & 0.808 $\pm$ 0.06 & 0.804 $\pm$ 0.02 \\

JacobiConv\cite{wang2022powerful} & -- & -- 
& 0.787 $\pm$ 0.02 & 0.814 $\pm$ 0.04 & 0.805 $\pm$ 0.06 & 0.807 $\pm$ 0.02 \\

\midrule

ARMA\cite{bianchi2021graph}& Baseline & -- 
& 0.758 $\pm$ 0.03 & 0.785 $\pm$ 0.03 & 0.779 $\pm$ 0.05 & 0.781 $\pm$ 0.03 \\

GAT\cite{velickovic2018graph} & Baseline & -- 
& 0.690 $\pm$ 0.06 & 0.705 $\pm$ 0.05 & 0.757 $\pm$ 0.10 & 0.729 $\pm$ 0.07 \\

\midrule

\textbf{ARMA} & \textbf{Mod\cite{tsitsulin2023graph}} & 0.01 
& \textbf{0.815 $\pm$ 0.01} & 0.854 $\pm$ 0.03 & 0.808 $\pm$ 0.04 & \textbf{0.829 $\pm$ 0.01} \\

GAT & Mod & -- 
& 0.781 $\pm$ 0.04 & 0.814 $\pm$ 0.06 & 0.805 $\pm$ 0.10 & 0.801 $\pm$ 0.05 \\

\textbf{ARMA} & \textbf{Cut\cite{bianchi2020spectral}} & 0.01 
& \textbf{0.815 $\pm$ 0.03} & \textbf{0.871 $\pm$ 0.04} & 0.786 $\pm$ 0.03 & 0.825 $\pm$ 0.03 \\

GAT & Cut & 0.9 
& 0.796 $\pm$ 0.03 & 0.861 $\pm$ 0.03 & 0.759 $\pm$ 0.07 & 0.804 $\pm$ 0.04 \\

\midrule

ARMA & DGI\cite{velickovic2019deep} + Mod & 0.01 
& 0.772 $\pm$ 0.02 & 0.789 $\pm$ 0.05 & 0.815 $\pm$ 0.05 & 0.799 $\pm$ 0.02 \\

GAT & DGI + Mod & 2 
& 0.755 $\pm$ 0.03 & 0.787 $\pm$ 0.05 & 0.776 $\pm$ 0.07 & 0.778 $\pm$ 0.03 \\

ARMA & DGI + Cut & 0.09 
& 0.782 $\pm$ 0.03 & 0.836 $\pm$ 0.05 & 0.762 $\pm$ 0.04 & 0.796 $\pm$ 0.03 \\

GAT & DGI + Cut & 0.9 
& 0.774 $\pm$ 0.03 & 0.816 $\pm$ 0.05 & 0.774 $\pm$ 0.05 & 0.792 $\pm$ 0.03 \\

\midrule

ARMA-C$^3$ & -- & 0.0001 
& 0.797 $\pm$ 0.02 & 0.819 $\pm$ 0.05 & \textbf{0.822 $\pm$ 0.05} & 0.817 $\pm$ 0.02 \\

GAT & MERIT\cite{jin2021multi} + Mod & 0.0001 
& 0.769 $\pm$ 0.03 & 0.793 $\pm$ 0.04 & 0.797 $\pm$ 0.07 & 0.792 $\pm$ 0.03 \\

ARMA & MERIT + Cut & 0.01 
& 0.787 $\pm$ 0.04 & 0.812 $\pm$ 0.04 & 0.807 $\pm$ 0.06 & 0.808 $\pm$ 0.03 \\

GAT & MERIT + Cut & 0.0001 
& 0.771 $\pm$ 0.03 & 0.794 $\pm$ 0.05 & 0.801 $\pm$ 0.06 & 0.795 $\pm$ 0.03 \\

\bottomrule
\end{tabular}
}
\end{adjustbox}
\end{subtable}

\bigskip

\begin{subtable}{0.97\textwidth}
\centering
\caption{semi-supervised classification (CN vs. AD, 10\% Training, 90\% Testing)}
\setlength\tabcolsep{5pt}
\begin{adjustbox}{max width=\textwidth}
{\small
\begin{tabular}{@{} l l c c c c c @{}}
\toprule
\textbf{Model} & \textbf{Loss Function} & $\boldsymbol{\lambda_{con}}$
& \textbf{Accuracy} & \textbf{Precision} & \textbf{Recall} & \textbf{F1 Score} \\
\midrule

SVC\cite{cortes1995support} & -- & -- 
& 0.749 $\pm$ 0.03 & 0.779 $\pm$ 0.05 & 0.532 $\pm$ 0.10 & 0.625 $\pm$ 0.07 \\

XGBoost\cite{chen2016xgboost} & -- & -- 
& 0.730 $\pm$ 0.03 & 0.696 $\pm$ 0.06 & 0.605 $\pm$ 0.10 & 0.640 $\pm$ 0.06 \\

MLPClassifier\cite{rumelhart1986learning} & -- & -- 
& 0.776 $\pm$ 0.03 & 0.726 $\pm$ 0.04 & 0.713 $\pm$ 0.06 & 0.718 $\pm$ 0.04 \\

Random Forest\cite{breiman2001random} & -- & -- 
& 0.758 $\pm$ 0.02 & 0.791 $\pm$ 0.04 & 0.551 $\pm$ 0.09 & 0.643 $\pm$ 0.06 \\

APPNP\cite{gasteiger2018predict} & -- & -- 
& 0.749 $\pm$ 0.05 & 0.707 $\pm$ 0.06 & 0.649 $\pm$ 0.15 & 0.667 $\pm$ 0.10 \\

JacobiConv\cite{wang2022powerful} & -- & -- 
& 0.762 $\pm$ 0.03 & 0.717 $\pm$ 0.06 & 0.694 $\pm$ 0.07 & 0.701 $\pm$ 0.03 \\

\midrule

ARMA~\cite{bianchi2021graph} & Baseline & -- 
& 0.739 $\pm$ 0.03 & 0.666 $\pm$ 0.04 & 0.710 $\pm$ 0.07 & 0.686 $\pm$ 0.04 \\

GAT\cite{velickovic2018graph} & Baseline & -- 
& 0.750 $\pm$ 0.06 & 0.711 $\pm$ 0.09 & 0.686 $\pm$ 0.11 & 0.689 $\pm$ 0.06 \\

\midrule

ARMA & Mod\cite{tsitsulin2023graph} & 0.01 
& 0.776 $\pm$ 0.04 & \textbf{0.817 $\pm$ 0.05} & 0.580 $\pm$ 0.13 & 0.668 $\pm$ 0.09 \\

\textbf{GAT} & \textbf{Mod} & 0.1 
& \textbf{0.791 $\pm$ 0.05} & 0.766 $\pm$ 0.09 & 0.717 $\pm$ 0.11 & 0.732 $\pm$ 0.06 \\

ARMA & Cut\cite{bianchi2020spectral}  & 0.01 
& 0.765 $\pm$ 0.04 & 0.666 $\pm$ 0.04 & \textbf{0.834 $\pm$ 0.05} & 0.740 $\pm$ 0.04 \\

GAT & Cut & 0.1 
& 0.778 $\pm$ 0.03 & 0.691 $\pm$ 0.05 & 0.824 $\pm$ 0.07 & \textbf{0.749 $\pm$ 0.04} \\

\midrule

ARMA & DGI\cite{velickovic2019deep} + Mod & 0.01 
& 0.759 $\pm$ 0.04 & 0.765 $\pm$ 0.06 & 0.595 $\pm$ 0.12 & 0.660 $\pm$ 0.08 \\

GAT & DGI + Mod & 1 
& 0.756 $\pm$ 0.04 & 0.699 $\pm$ 0.06 & 0.714 $\pm$ 0.06 & 0.703 $\pm$ 0.04 \\

ARMA & DGI + Cut & 0.01 
& 0.782 $\pm$ 0.03 & 0.739 $\pm$ 0.06 & 0.725 $\pm$ 0.11 & 0.724 $\pm$ 0.06 \\

GAT & DGI + Cut & 1 
& 0.763 $\pm$ 0.04 & 0.690 $\pm$ 0.06 & 0.765 $\pm$ 0.05 & 0.723 $\pm$ 0.04 \\

\midrule

ARMA-C$^3$ & -- & 0.1 
& 0.779 $\pm$ 0.04 & 0.728 $\pm$ 0.07 & 0.719 $\pm$ 0.13 & 0.713 $\pm$ 0.06 \\

GAT & MERIT\cite{jin2021multi} + Mod & 0.1 
& 0.763 $\pm$ 0.03 & 0.699 $\pm$ 0.04 & 0.727 $\pm$ 0.10 & 0.709 $\pm$ 0.05 \\

ARMA & MERIT + Cut & 0.01 
& 0.762 $\pm$ 0.03 & 0.759 $\pm$ 0.05 & 0.612 $\pm$ 0.12 & 0.667 $\pm$ 0.08 \\

GAT & MERIT + Cut & 0.01 
& 0.787 $\pm$ 0.03 & 0.729 $\pm$ 0.03 & 0.755 $\pm$ 0.09 & 0.738 $\pm$ 0.05 \\

\bottomrule
\end{tabular}
}
\end{adjustbox}
\end{subtable}

\end{table*}
\begin{table*}[!htbp]
\centering
\caption{Comparison of semi-supervised classification performance on MedMNIST datasets.
Metrics are reported as mean $\pm$ standard deviation over 20-fold cross-validation.}
\label{tab:medmnist_combined}

\begin{subtable}{0.97\textwidth}
\centering
\caption{semi-supervised classification on BreastMNIST, 10\% Training, 90\% Testing}
\setlength\tabcolsep{5pt}
\begin{adjustbox}{max width=\textwidth}
{\small
\begin{tabular}{@{} l l c c c c c @{}}
\toprule
\textbf{Model} & \textbf{Loss Function} & $\boldsymbol{\lambda_{con}}$
& \textbf{Accuracy} & \textbf{Precision} & \textbf{Recall} & \textbf{F1 Score} \\
\midrule

SVC\cite{cortes1995support} & -- & -- 
& 0.839 $\pm$ 0.01 & 0.836 $\pm$ 0.02 & 0.972 $\pm$ 0.01 & 0.899 $\pm$ 0.01 \\

XGBoost\cite{chen2016xgboost} & -- & -- 
& 0.809 $\pm$ 0.02 & 0.825 $\pm$ 0.01 & 0.937 $\pm$ 0.02 & 0.877 $\pm$ 0.01 \\

MLPClassifier\cite{rumelhart1986learning} & -- & -- 
& 0.830 $\pm$ 0.02 & 0.897 $\pm$ 0.02 & 0.868 $\pm$ 0.02 & 0.882 $\pm$ 0.01 \\

Random Forest\cite{breiman2001random} & -- & -- 
& 0.798 $\pm$ 0.02 & 0.791 $\pm$ 0.02 & \textbf{0.985 $\pm$ 0.01} & 0.877 $\pm$ 0.01 \\

APPNP\cite{gasteiger2018predict} & -- & N/A 
& 0.841 $\pm$ 0.02 & 0.867 $\pm$ 0.02 & 0.926 $\pm$ 0.03 & 0.895 $\pm$ 0.01 \\

JacobiConv\cite{wang2022powerful} & -- & N/A 
& 0.817 $\pm$ 0.03 & 0.862 $\pm$ 0.05 & 0.900 $\pm$ 0.07 & 0.877 $\pm$ 0.03 \\

\midrule

ARMA\cite{bianchi2021graph} & Baseline & N/A 
& 0.846 $\pm$ 0.02 & 0.876 $\pm$ 0.02 & 0.920 $\pm$ 0.03 & 0.897 $\pm$ 0.01 \\

GAT\cite{velickovic2018graph} & Baseline & N/A 
& 0.740 $\pm$ 0.05 & 0.858 $\pm$ 0.03 & 0.777 $\pm$ 0.12 & 0.809 $\pm$ 0.06 \\

\midrule

ARMA & Mod\cite{tsitsulin2023graph} & 0.001 
& 0.843 $\pm$ 0.02 & 0.873 $\pm$ 0.02 & 0.921 $\pm$ 0.03 & 0.896 $\pm$ 0.01 \\

GAT & Mod & 0.001 
& 0.754 $\pm$ 0.03 & 0.866 $\pm$ 0.03 & 0.788 $\pm$ 0.06 & 0.823 $\pm$ 0.03 \\

ARMA & Cut\cite{bianchi2020spectral} & 0.001 
& 0.844 $\pm$ 0.02 & 0.877 $\pm$ 0.02 & 0.916 $\pm$ 0.03 & 0.896 $\pm$ 0.02 \\

GAT & Cut & 0.0001 
& 0.725 $\pm$ 0.07 & 0.877 $\pm$ 0.04 & 0.734 $\pm$ 0.13 & 0.788 $\pm$ 0.10 \\

\midrule

ARMA & DGI\cite{velickovic2019deep} + Mod & 0.01 
& 0.790 $\pm$ 0.03 & 0.897 $\pm$ 0.02 & 0.806 $\pm$ 0.04 & 0.848 $\pm$ 0.02 \\

ARMA & DGI + Cut & 0.01 
& 0.730 $\pm$ 0.04 & \textbf{0.908 $\pm$ 0.02} & 0.704 $\pm$ 0.08 & 0.790 $\pm$ 0.04 \\

GAT & DGI + Mod & 0.01 
& 0.695 $\pm$ 0.06 & 0.888 $\pm$ 0.03 & 0.668 $\pm$ 0.09 & 0.758 $\pm$ 0.06 \\

GAT & DGI + Cut & 0.001 
& 0.677 $\pm$ 0.06 & 0.876 $\pm$ 0.03 & 0.654 $\pm$ 0.11 & 0.742 $\pm$ 0.06 \\

\midrule

ARMA-C$^3$ & -- & 0.0001 
& 0.855 $\pm$ 0.02 & 0.890 $\pm$ 0.02 & 0.917 $\pm$ 0.03 & 0.902 $\pm$ 0.01 \\

GAT & MERIT\cite{jin2021multi} + Mod & 0.0001 
& 0.751 $\pm$ 0.05 & 0.899 $\pm$ 0.02 & 0.744 $\pm$ 0.09 & 0.810 $\pm$ 0.05 \\

\textbf{ARMA} & \textbf{MERIT + Cut} & 0.0001 
& \textbf{0.857 $\pm$ 0.02} & 0.892 $\pm$ 0.02 & 0.916 $\pm$ 0.03 & \textbf{0.903 $\pm$ 0.01} \\

GAT & MERIT + Cut & 0.0001 
& 0.753 $\pm$ 0.06 & 0.898 $\pm$ 0.02 & 0.750 $\pm$ 0.11 & 0.812 $\pm$ 0.06 \\

\bottomrule
\end{tabular}
}
\end{adjustbox}
\end{subtable}

\bigskip

\begin{subtable}{0.97\textwidth}
\centering
\caption{semi-supervised classification on PneumoniaMNIST, 10\% Training, 90\% Testing}
\setlength\tabcolsep{5pt}
\begin{adjustbox}{max width=\textwidth}
{\small
\begin{tabular}{@{} l l c c c c c @{}}
\toprule
\textbf{Model} & \textbf{Loss Function} & $\boldsymbol{\lambda_{con}}$
& \textbf{Accuracy} & \textbf{Precision} & \textbf{Recall} & \textbf{F1 Score} \\
\midrule

SVC\cite{cortes1995support} & -- & -- 
& 0.939 $\pm$ 0.01 & 0.950 $\pm$ 0.01 & 0.943 $\pm$ 0.01 & 0.946 $\pm$ 0.00 \\

XGBoost\cite{chen2016xgboost} & -- & -- 
& 0.927 $\pm$ 0.01 & 0.937 $\pm$ 0.01 & 0.933 $\pm$ 0.01 & 0.935 $\pm$ 0.00 \\

MLPClassifier\cite{rumelhart1986learning} & -- & -- 
& 0.938 $\pm$ 0.01 & 0.952 $\pm$ 0.01 & 0.937 $\pm$ 0.01 & 0.944 $\pm$ 0.00 \\

Random Forest\cite{breiman2001random} & -- & -- 
& 0.927 $\pm$ 0.00 & 0.938 $\pm$ 0.01 & 0.932 $\pm$ 0.01 & 0.935 $\pm$ 0.00 \\

APPNP\cite{gasteiger2018predict} & -- & N/A 
& 0.940 $\pm$ 0.00 & 0.954 $\pm$ 0.01 & 0.938 $\pm$ 0.01 & 0.946 $\pm$ 0.00 \\

JacobiConv\cite{wang2022powerful} & -- & N/A 
& 0.929 $\pm$ 0.01 & 0.940 $\pm$ 0.03 & 0.934 $\pm$ 0.02 & 0.936 $\pm$ 0.01 \\

\midrule

ARMA\cite{bianchi2021graph} & Baseline & N/A 
& 0.939 $\pm$ 0.01 & 0.954 $\pm$ 0.01 & 0.937 $\pm$ 0.01 & 0.945 $\pm$ 0.00 \\

GAT\cite{velickovic2018graph} & Baseline & N/A 
& 0.931 $\pm$ 0.00 & 0.944 $\pm$ 0.01 & 0.931 $\pm$ 0.02 & 0.937 $\pm$ 0.00 \\

\midrule

ARMA & Mod\cite{tsitsulin2023graph} & 0.1 
& 0.940 $\pm$ 0.00 & 0.956 $\pm$ 0.01 & 0.936 $\pm$ 0.01 & 0.946 $\pm$ 0.00 \\

\textbf{GAT} & \textbf{Cut} & 0.1 
& 0.913 $\pm$ 0.00 & \textbf{0.978 $\pm$ 0.00} & 0.863 $\pm$ 0.00 & 0.917 $\pm$ 0.00 \\

ARMA & Cut\cite{bianchi2020spectral} & 0.1 
& 0.934 $\pm$ 0.01 & 0.971 $\pm$ 0.01 & 0.908 $\pm$ 0.01 & 0.939 $\pm$ 0.01 \\

GAT & Mod & 0.1 
& 0.933 $\pm$ 0.00 & 0.960 $\pm$ 0.01 & 0.918 $\pm$ 0.01 & 0.939 $\pm$ 0.00 \\

\midrule

ARMA & DGI\cite{velickovic2019deep} + Mod & 0.01 
& 0.936 $\pm$ 0.01 & 0.942 $\pm$ 0.02 & 0.943 $\pm$ 0.02 & 0.942 $\pm$ 0.01 \\

GAT & DGI + Mod & 1 
& 0.882 $\pm$ 0.03 & 0.977 $\pm$ 0.01 & 0.807 $\pm$ 0.07 & 0.883 $\pm$ 0.04 \\

ARMA & DGI + Cut & 0.01 
& 0.937 $\pm$ 0.01 & 0.945 $\pm$ 0.02 & 0.942 $\pm$ 0.01 & 0.943 $\pm$ 0.01 \\

GAT & DGI + Cut & 1 
& 0.890 $\pm$ 0.03 & 0.975 $\pm$ 0.02 & 0.825 $\pm$ 0.07 & 0.892 $\pm$ 0.04 \\

\midrule

ARMA-C$^3$ & -- & 0.005 
& \textbf{0.944 $\pm$ 0.01} & 0.952 $\pm$ 0.01 & \textbf{0.948 $\pm$ 0.01} & \textbf{0.950 $\pm$ 0.01} \\

GAT & MERIT\cite{jin2021multi} + Mod & 0.005 
& 0.932 $\pm$ 0.01 & 0.948 $\pm$ 0.02 & 0.931 $\pm$ 0.02 & 0.939 $\pm$ 0.01 \\

\textbf{ARMA} & \textbf{MERIT + Cut} & 0.005 
& \textbf{0.944 $\pm$ 0.01} & 0.952 $\pm$ 0.01 & \textbf{0.948 $\pm$ 0.01} & \textbf{0.950 $\pm$ 0.00} \\

GAT & MERIT + Cut & 0.005 
& 0.933 $\pm$ 0.01 & 0.946 $\pm$ 0.02 & 0.934 $\pm$ 0.02 & 0.940 $\pm$ 0.01 \\

\bottomrule
\end{tabular}
}
\end{adjustbox}
\end{subtable}
\end{table*}
\begin{table*}[t]
\centering
\caption{Fully supervised evaluation of ARMA-C$^3$ under complete label availability using an 80/20 train-test split. Results are reported as mean $\pm$ standard deviation across ten repeated runs.}
\label{tab:fully_supervised_results}
\resizebox{\textwidth}{!}{
\begin{tabular}{lcccc}
\toprule
\textbf{Dataset} & \textbf{Accuracy} & \textbf{Precision} & \textbf{Recall} & \textbf{F1-score} \\
\midrule
ADNI (CN vs. MCI) 
& 0.951 $\pm$ 0.01
& 0.947 $\pm$ 0.02
& 0.966 $\pm$ 0.01
& 0.956 $\pm$ 0.01 \\

ADNI (CN vs. AD)
& 0.910 $\pm$ 0.03
& 0.900 $\pm$ 0.05
& 0.875 $\pm$ 0.07
& 0.885 $\pm$ 0.04 \\

NIFD
& 0.925 $\pm$ 0.01
& 0.933 $\pm$ 0.02
& 0.958 $\pm$ 0.02
& 0.945 $\pm$ 0.01 \\

BreastMNIST
& 0.978 $\pm$ 0.01
& 0.985 $\pm$ 0.00
& 0.985 $\pm$ 0.01
& 0.985 $\pm$ 0.00 \\

PneumoniaMNIST
& 0.984 $\pm$ 0.00
& 0.988 $\pm$ 0.00
& 0.984 $\pm$ 0.00
& 0.986 $\pm$ 0.00 \\

Liver Ultrasound
& 0.964 $\pm$ 0.01
& 0.969 $\pm$ 0.01
& 0.979 $\pm$ 0.01
& 0.974 $\pm$ 0.00 \\
\bottomrule
\end{tabular}
}
\end{table*}

In the semi-supervised setting, ARMA-C$^3$ consistently improves liver ultrasound classification performance (Table~\ref{tab:liver_unsup_results}(b)) under limited label availability, highlighting the effectiveness of graph-based contrastive representation learning.
On BreastMNIST and PneumoniaMNIST (Table~\ref{tab:medmnist_combined}), ARMA-C$^3$ maintains superior performance over conventional classifiers. MERIT-based variants consistently yield the highest accuracy and F1-scores, demonstrating that contrastive consistency along with regularization remain beneficial even in semi-supervised settings.
In summary, DGI works well in purely unsupervised scenarios due to its global information preservation, while MERIT is better suited for semi-supervised learning as it effectively leverages sparse labels through stable teacher–student alignment.

To contextualize the semi-supervised and unsupervised results, we additionally evaluate ARMA-C$^3$ under full label availability using an 80/20 train-test split as an upper-bound reference (Table~\ref{tab:fully_supervised_results}). The framework retains its joint structural regularization and contrastive learning objectives, achieving strong and stable performance across all datasets, including F1-scores of 0.956 (CN vs. MCI), 0.885 (CN vs. AD), 0.945 (NIFD), 0.985 (BreastMNIST), and 0.974 (liver ultrasound). These results indicate that ARMA-C$^3$ remains effective across different supervision regimes while preserving robust representation learning across heterogeneous biomedical imaging modalities.

Overall, these results show that ARMA-C$^3$ performs robustly in both unsupervised and semi-supervised scenarios, making it well-suited for biomedical applications where labeled data are scarce.

\section{Ablation Study}
\label{sec:ablation}  
To examine the role of each part in the ARMA-C$^3$ framework, we perform thorough ablation studies on all datasets. For each dataset, we change key design choices, such as feature representation, graph construction, architectural parts, and contrastive regularization. Unless stated otherwise, all ablations take place in an unsupervised setting using the ARMA backbone, along with modularity-based structural regularization and MERIT contrastive learning. We report performance as the mean $\pm$ standard deviation over ten independent runs.
\subsection{ADNI (Diffusion MRI): CN vs.\ MCI}
\paragraph{Effect of histogram bin resolution:} We first look at how feature granularity affects results by lowering the resolution of Fractional Anisotropy (FA) histograms from the default 20 bins to 10 bins. With the 10-bin FA histograms, ARMA-C$^3$ achieves an accuracy of $0.7124 \pm 0.0016$, a precision of $0.7685 \pm 0.0028$, a recall of $0.6978 \pm 0.0045$, and an F1-score of $0.7298 \pm 0.0020$. This change leads to a steady drop in performance compared to the 20-bin setup. This shows that finer histogram detail is crucial for maintaining subtle microstructural differences between CN and MCI subjects.
\paragraph{Effect of feature modality:}  
We evaluate whether adding Axial Diffusivity (AD) with FA improves discriminative performance. Surprisingly, combining FA and AD features leads to lower accuracy and much higher variance. The results show an accuracy of $0.6029 \pm 0.0735$, precision of $0.6731 \pm 0.0976$, recall of $0.7375 \pm 0.1824$, and an F1-score of $0.6547 \pm 0.0828$.  

These findings suggest that AD brings more variability, which weakens the similarity structure of graphs in early-stage disease classification. 
\paragraph{Effect of graph sparsification threshold $\alpha$:}
The threshold $\alpha$ determines graph connectivity by controlling which
pairwise feature similarities are retained as edges. Table~\ref{tab:ablation_alpha_cn_mci}
reports the clustering performance across different $\alpha$ values.
\begin{table}[!htbp]
\centering
\caption{Ablation study on the effect of graph sparsification threshold $\alpha$
for CN vs.\ MCI classification on the ADNI diffusion MRI dataset.
Results are reported as mean $\pm$ standard deviation over ten runs.}
\label{tab:ablation_alpha_cn_mci}
\setlength{\tabcolsep}{6pt}
\renewcommand{\arraystretch}{1.05}
\begin{tabular}{c c c c c}
\toprule
$\alpha$ & Accuracy & Precision & Recall & F1 Score \\
\midrule
0.20 & 0.757 $\pm$ 0.02 & 0.815 $\pm$ 0.02 & 0.735 $\pm$ 0.01 & 0.772 $\pm$ 0.02 \\
0.40 & 0.760 $\pm$ 0.02 & 0.818 $\pm$ 0.02 & 0.739 $\pm$ 0.02 & 0.775 $\pm$ 0.02 \\
0.60 & 0.760 $\pm$ 0.02 & 0.819 $\pm$ 0.02 & 0.737 $\pm$ 0.01 & 0.774 $\pm$ 0.02 \\
0.85 & 0.726 $\pm$ 0.00 & 0.772 $\pm$ 0.00 & 0.726 $\pm$ 0.00 & 0.743 $\pm$ 0.00 \\
0.92 & 0.774 $\pm$ 0.01 & 0.835 $\pm$ 0.01 & 0.740 $\pm$ 0.00 & 0.785 $\pm$ 0.01 \\
0.95 & 0.703 $\pm$ 0.00 & 0.757 $\pm$ 0.01 & 0.692 $\pm$ 0.00 & 0.722 $\pm$ 0.00 \\
\bottomrule
\end{tabular}
\end{table}
\paragraph{Effect of activation function:}
We investigate the impact of different nonlinear activation functions within the
ARMA encoder. As shown in Table~\ref{tab:ablation_activ_cn_mci}, ReLU and SELU
produce highly comparable results, with only marginal differences in accuracy
and F1-score. This indicates that ARMA-C$^3$ is robust to the choice of
activation function in this setting.
\begin{table}[!htbp]
\centering
\caption{Ablation study on the effect of activation function in the ARMA encoder
for CN vs.\ MCI classification on the ADNI diffusion MRI dataset.
Results are reported as mean $\pm$ standard deviation over ten runs.}
\label{tab:ablation_activ_cn_mci}
\setlength{\tabcolsep}{6pt}
\renewcommand{\arraystretch}{1.05}
\begin{tabular}{l c c c c}
\toprule
Activation & Accuracy & Precision & Recall & F1 Score \\
\midrule
ReLU & 0.773 $\pm$ 0.00 & 0.834 $\pm$ 0.01 & 0.746 $\pm$ 0.00 & 0.786 $\pm$ 0.00 \\
SELU & 0.772 $\pm$ 0.00 & 0.834 $\pm$ 0.00 & 0.748 $\pm$ 0.00 & 0.787 $\pm$ 0.00 \\
ELU & 0.774 $\pm$ 0.01 & 0.835 $\pm$ 0.01 & 0.740 $\pm$ 0.00 & 0.785 $\pm$ 0.01 \\ 
\bottomrule
\end{tabular}
\end{table}
\begin{table}[!htbp]
\centering
\caption{Ablation study on the effect of contrastive loss weight
$\lambda_{\text{con}}$ for CN vs.\ MCI classification on the ADNI diffusion MRI
dataset. Results are reported as mean $\pm$ standard deviation over ten runs.}
\label{tab:ablation_lambda_cn_mci}
\setlength{\tabcolsep}{6pt}
\renewcommand{\arraystretch}{1.05}
\begin{tabular}{c c c c c}
\toprule
$\lambda_{\text{con}}$ & Accuracy & Precision & Recall & F1 Score \\
\midrule
0.001 & 0.761 $\pm$ 0.01 & 0.818 $\pm$ 0.01 & 0.740 $\pm$ 0.01 & 0.776 $\pm$ 0.01 \\
0.01  & 0.762 $\pm$ 0.01 & 0.819 $\pm$ 0.01 & 0.741 $\pm$ 0.01 & 0.776 $\pm$ 0.01 \\
0.3   & 0.773 $\pm$ 0.00 & 0.834 $\pm$ 0.01 & 0.746 $\pm$ 0.00 & 0.786 $\pm$ 0.00 \\
0.9   & 0.773 $\pm$ 0.00 & 0.834 $\pm$ 0.01 & 0.745 $\pm$ 0.00 & 0.785 $\pm$ 0.00 \\
2.0   & 0.772 $\pm$ 0.00 & 0.833 $\pm$ 0.01 & 0.745 $\pm$ 0.00 & 0.785 $\pm$ 0.00 \\
\bottomrule
\end{tabular}
\end{table}
\paragraph{Effect of contrastive loss weight $\lambda_{\text{con}}$:}
We examine how the contrastive loss weight, $\lambda_{\text{con}}$, affects clustering performance. The findings in Table~\ref{tab:ablation_lambda_cn_mci} show that a moderate level of contrastive regularization, around $\lambda_{\text{con}} \approx 0.3$, offers the best balance between representation invariance and structural fidelity. In contrast, both weaker and stronger regularization result in slightly worse outcomes.
\subsection{ADNI (Diffusion MRI): CN vs.\ AD}
\paragraph{Effect of histogram bin resolution:} We evaluate how the proposed framework reacts to changes in histogram granularity in the CN vs. AD setting by reducing the number of FA histogram bins from the default 20 to 10. With the 10-bin FA histograms, ARMA-C$^3$ achieves an accuracy of $0.6697 \pm 0.02$, a precision of $0.5714 \pm 0.02$, a recall of $0.6818 \pm 0.03$, and an F1-score of $0.6218 \pm 0.02$. This reduction in histogram resolution leads to lower performance compared to the default setup. It indicates that having more detailed FA distributions is important for showing disease-related microstructural differences between CN and AD subjects.
\paragraph{Effect of feature modality:} We also check whether combining Axial Diffusivity (AD) with FA improves performance in the CN vs. AD setting. However, merging FA and AD features results in lower performance, with an accuracy of $0.5158 \pm 0.03$, precision of $0.4296 \pm 0.02$, recall of $0.6591 \pm 0.03$, and an F1-score of $0.5202 \pm 0.02$. These findings suggest that AD adds extra variability that disrupts the similarity structure needed for stable graph-based learning.
\paragraph{Effect of graph sparsification threshold $\alpha$:}
Table~\ref{tab:ablation_alpha_cn_ad} reports the effect of varying $\alpha$ for
CN vs.\ AD clustering. Performance remains relatively stable for moderate
thresholds, while higher thresholds degrade recall due to excessive graph
sparsification. Compared to CN vs.\ MCI, the CN vs.\ AD task is less sensitive
to $\alpha$, reflecting clearer disease separation.
\begin{table}[!htbp]
\centering
\caption{Ablation study on the effect of graph sparsification threshold $\alpha$
for CN vs.\ AD classification on the ADNI diffusion MRI dataset.
Results are reported as mean $\pm$ standard deviation over ten runs.}
\label{tab:ablation_alpha_cn_ad}
\setlength{\tabcolsep}{6pt}
\renewcommand{\arraystretch}{1.05}
\begin{tabular}{c c c c c}
\toprule
$\alpha$ & Accuracy & Precision & Recall & F1 Score \\
\midrule
0.20 & 0.748 $\pm$ 0.02 & 0.651 $\pm$ 0.02 & 0.780 $\pm$ 0.02 & 0.708 $\pm$ 0.02 \\
0.60 & 0.749 $\pm$ 0.02 & 0.653 $\pm$ 0.03 & 0.783 $\pm$ 0.03 & 0.709 $\pm$ 0.03 \\
0.75 & 0.745 $\pm$ 0.01 & 0.644 $\pm$ 0.01 & 0.798 $\pm$ 0.01 & 0.710 $\pm$ 0.01 \\
0.80 & 0.748 $\pm$ 0.01 & 0.642 $\pm$ 0.01 & 0.805 $\pm$ 0.01 & 0.712 $\pm$ 0.00 \\
0.85 & 0.732 $\pm$ 0.00 & 0.628 $\pm$ 0.01 & 0.794 $\pm$ 0.01 & 0.699 $\pm$ 0.01 \\
0.90 & 0.735 $\pm$ 0.00 & 0.637 $\pm$ 0.01 & 0.772 $\pm$ 0.01 & 0.695 $\pm$ 0.00 \\
\bottomrule
\end{tabular}
\end{table}

\paragraph{Effect of activation function:}
As shown in Table~\ref{tab:ablation_activ_cn_ad}, ReLU and SELU again yield
nearly identical performance for CN vs.\ AD classification. This further
confirms that ARMA-C$^3$ does not rely on a specific activation choice and
exhibits consistent behavior across disease severity levels.
\begin{table}[!htbp]
\centering
\caption{Ablation study on the effect of activation function in the ARMA encoder
for CN vs.\ AD classification on the ADNI diffusion MRI dataset.
Results are reported as mean $\pm$ standard deviation over ten runs.}
\label{tab:ablation_activ_cn_ad}
\setlength{\tabcolsep}{6pt}
\renewcommand{\arraystretch}{1.05}
\begin{tabular}{l c c c c}
\toprule
Activation & Accuracy & Precision & Recall & F1 Score \\
\midrule
ReLU & 0.746 $\pm$ 0.00 & 0.643 $\pm$ 0.01 & 0.805 $\pm$ 0.01 & 0.712 $\pm$ 0.01 \\
SELU & 0.744 $\pm$ 0.01 & 0.643 $\pm$ 0.01 & 0.809 $\pm$ 0.01 & 0.711 $\pm$ 0.01 \\
ELU & 0.748 $\pm$ 0.01 & 0.642 $\pm$ 0.01 & 0.805 $\pm$ 0.01 & 0.712 $\pm$ 0.00 \\
\bottomrule
\end{tabular}
\end{table}
\paragraph{Effect of contrastive loss weight $\lambda_{\text{con}}$:}
Table~\ref{tab:ablation_lambda_cn_ad} summarizes the influence of
$\lambda_{\text{con}}$ on CN vs.\ AD performance. We observe similar results by significantly varying the $\lambda_{\text{con}}$ values.
\begin{table}[!htbp]
\centering
\caption{Ablation study on the effect of contrastive loss weight
$\lambda_{\text{con}}$ for CN vs.\ AD classification on the ADNI diffusion MRI
dataset. Results are reported as mean $\pm$ standard deviation over ten runs.}
\label{tab:ablation_lambda_cn_ad}
\setlength{\tabcolsep}{6pt}
\renewcommand{\arraystretch}{1.05}
\begin{tabular}{c c c c c}
\toprule
$\lambda_{\text{con}}$ & Accuracy & Precision & Recall & F1 Score \\
\midrule
0.001 & 0.746 $\pm$ 0.01 & 0.645 $\pm$ 0.01 & 0.799 $\pm$ 0.01 & 0.711 $\pm$ 0.01 \\
0.01  & 0.747 $\pm$ 0.01 & 0.645 $\pm$ 0.01 & 0.800 $\pm$ 0.01 & 0.712 $\pm$ 0.01 \\
0.5   & 0.748 $\pm$ 0.00 & 0.643 $\pm$ 0.01 & 0.805 $\pm$ 0.01 & 0.712 $\pm$ 0.01 \\
0.9   & 0.745 $\pm$ 0.00 & 0.642 $\pm$ 0.01 & 0.804 $\pm$ 0.01 & 0.711 $\pm$ 0.01 \\
2.0   & 0.744 $\pm$ 0.00 & 0.641 $\pm$ 0.01 & 0.803 $\pm$ 0.01 & 0.710 $\pm$ 0.01 \\
\bottomrule
\end{tabular}
\end{table}
\subsection{NIFD (Diffusion MRI)}

\paragraph{Effect of activation function:}
The activation function influences non-linear feature propagation and representation learning in the ARMA encoder. As shown in Table ~\ref{tab:ablation_activ_nifd}, ELU achieves the best overall performance on the NIFD diffusion MRI dataset.
\begin{table}[!htbp]
\centering
\caption{Ablation study on the effect of activation function in the ARMA encoder
for the NIFD diffusion MRI dataset. Results are reported as mean $\pm$ standard
deviation over ten runs.}
\label{tab:ablation_activ_nifd}
\setlength{\tabcolsep}{6pt}
\renewcommand{\arraystretch}{1.05}
\begin{tabular}{l c c c c}
\toprule
Activation & Accuracy & Precision & Recall & F1 Score \\
\midrule
SELU & 0.695 $\pm$ 0.01 & 0.882 $\pm$ 0.02 & 0.631 $\pm$ 0.01 & 0.735 $\pm$ 0.01 \\
ReLU & 0.701 $\pm$ 0.01 & 0.887 $\pm$ 0.01 & 0.637 $\pm$ 0.01 & 0.741 $\pm$ 0.01 \\
ELU & 0.712 $\pm$ 0.01 & 0.887 $\pm$ 0.01 & 0.638 $\pm$ 0.01 & 0.742 $\pm$ 0.01 \\
\bottomrule
\end{tabular}
\end{table}
\begin{table}[!htbp]
\centering
\caption{Ablation study on the effect of graph sparsification threshold $\alpha$
for the NIFD diffusion MRI dataset. Results are reported as mean $\pm$ standard
deviation over ten runs.}
\label{tab:ablation_alpha_nifd}
\setlength{\tabcolsep}{6pt}
\renewcommand{\arraystretch}{1.05}
\begin{tabular}{c c c c c}
\toprule
$\alpha$ & Accuracy & Precision & Recall & F1 Score \\
\midrule
0.20 & 0.607 $\pm$ 0.07 & 0.771 $\pm$ 0.06 & 0.587 $\pm$ 0.06 & 0.666 $\pm$ 0.06 \\
0.5 & 0.712 $\pm$ 0.01 & 0.887 $\pm$ 0.01 & 0.638 $\pm$ 0.01 & 0.742 $\pm$ 0.01 \\
0.55 & 0.656 $\pm$ 0.01 & 0.851 $\pm$ 0.01 & 0.592 $\pm$ 0.01 & 0.698 $\pm$ 0.01 \\
0.60 & 0.573 $\pm$ 0.01 & 0.763 $\pm$ 0.02 & 0.528 $\pm$ 0.01 & 0.624 $\pm$ 0.01 \\
0.80 & 0.521 $\pm$ 0.04 & 0.688 $\pm$ 0.05 & 0.526 $\pm$ 0.02 & 0.596 $\pm$ 0.03 \\
0.90 & 0.510 $\pm$ 0.00 & 0.681 $\pm$ 0.01 & 0.507 $\pm$ 0.02 & 0.581 $\pm$ 0.01 \\
\bottomrule
\end{tabular}
\end{table}

\paragraph{Effect of graph sparsification threshold $\alpha$:} 
The graph sparsification threshold $\alpha$ controls the graph connectivity by filtering weak edges, hence affecting the information propagation and clustering quality. As shown in Table~\ref{tab:ablation_alpha_nifd}, the best performance is achieved with moderate sparsification ($\alpha=0.5$), whereas graphs that are too sparse degrade both representation learning and clustering consistency.
\paragraph{Effect of contrastive loss weight $\lambda_{\text{con}}$:}
The contrastive loss weight $\lambda_{\text{con}}$ controls the trade-off between the structural regularization and contrastive representation learning. Table~\ref{tab:ablation_lambda_nifd} shows that moderate contrastive weighting yields the best performance, suggesting the importance of balancing graph structure preservation and representation consistency.
\begin{table}[!htbp]
\centering
\caption{Ablation study on the effect of contrastive loss weight
$\lambda_{\text{con}}$ for the NIFD diffusion MRI dataset. Results are reported
as mean $\pm$ standard deviation over ten runs.}
\label{tab:ablation_lambda_nifd}
\setlength{\tabcolsep}{6pt}
\renewcommand{\arraystretch}{1.05}
\begin{tabular}{c c c c c}
\toprule
$\lambda_{\text{con}}$ & Accuracy & Precision & Recall & F1 Score \\
\midrule
0.001 & 0.700 $\pm$ 0.02 & 0.894 $\pm$ 0.02 & 0.628 $\pm$ 0.02 & 0.737 $\pm$ 0.02 \\
0.01  & 0.712 $\pm$ 0.01 & 0.887 $\pm$ 0.01 & 0.638 $\pm$ 0.01 & 0.742 $\pm$ 0.01 \\
0.09  & 0.700 $\pm$ 0.01 & 0.884 $\pm$ 0.01 & 0.637 $\pm$ 0.01 & 0.740 $\pm$ 0.01 \\
0.3   & 0.713 $\pm$ 0.01 & 0.887 $\pm$ 0.01 & 0.639 $\pm$ 0.01 & 0.743 $\pm$ 0.01 \\
2.0   & 0.701 $\pm$ 0.01 & 0.886 $\pm$ 0.01 & 0.636 $\pm$ 0.01 & 0.740 $\pm$ 0.01 \\
5.0   & 0.701 $\pm$ 0.01 & 0.887 $\pm$ 0.01 & 0.637 $\pm$ 0.01 & 0.741 $\pm$ 0.01 \\
\bottomrule
\end{tabular}
\end{table}

\subsection{BreastMNIST}
\paragraph{Effect of feature backbone:}
The backbone used for feature extraction has a significant impact on representation quality and clustering performance. As shown in Table~\ref{tab:ablation_backbone_breastmnist}, ViT-DINO significantly outperforms ResNet-18 across all evaluation metrics. These results highlight the importance of strong pretrained feature representations in biomedical graph learning. The superior performance of ViT-DINO suggests that transformer-based self-supervised embeddings capture more discriminative and semantically meaningful structural patterns than conventional CNN-based features, leading to improved graph construction and cluster separability. 
\begin{table}[!htbp]
\centering
\caption{Ablation study on the effect of feature extraction backbone for the
BreastMNIST dataset. Results are reported as mean $\pm$ standard deviation over
ten runs.}
\label{tab:ablation_backbone_breastmnist}
\setlength{\tabcolsep}{6pt}
\renewcommand{\arraystretch}{1.05}
\begin{tabular}{l c c c c}
\toprule
Backbone & Accuracy & Precision & Recall & F1 Score \\
\midrule
ResNet-18 & 0.619 $\pm$ 0.00 & 0.704 $\pm$ 0.00 & 0.610 $\pm$ 0.00 & 0.654 $\pm$ 0.00 \\
ViT-DINO & 0.731 $\pm$ 0.00 & 0.731 $\pm$ 0.00 & 1.000 $\pm$ 0.00 & 0.844 $\pm$ 0.00 \\
\bottomrule
\end{tabular}
\end{table}
\paragraph{Effect of graph sparsification threshold $\alpha$:}
We then study the sensitivity of the BreastMNIST clustering performance to the graph sparsification threshold $\alpha$ as shown in Table~\ref{tab:ablation_alpha_breastmnist}.
\begin{table}[!htbp]
\centering
\caption{Ablation study on the effect of graph sparsification threshold $\alpha$
for the BreastMNIST dataset. Results are reported as mean $\pm$ standard
deviation over ten runs.}
\label{tab:ablation_alpha_breastmnist}
\setlength{\tabcolsep}{6pt}
\renewcommand{\arraystretch}{1.05}
\begin{tabular}{c c c c c}
\toprule
$\alpha$ & Accuracy & Precision & Recall & F1 Score \\
\midrule
0.20 & 0.725 $\pm$ 0.01 & 0.733 $\pm$ 0.00 & 0.982 $\pm$ 0.01 & 0.839 $\pm$ 0.00 \\
0.40 & 0.721 $\pm$ 0.00 & 0.721 $\pm$ 0.00 & 1.000 $\pm$ 0.00 & 0.844 $\pm$ 0.00 \\
0.73 & 0.731 $\pm$ 0.00 & 0.731 $\pm$ 0.00 & 0.979 $\pm$ 0.01 & 0.844 $\pm$ 0.00 \\
0.75 & 0.716 $\pm$ 0.01 & 0.731 $\pm$ 0.00 & 0.967 $\pm$ 0.01 & 0.832 $\pm$ 0.01 \\
0.80 & 0.701 $\pm$ 0.01 & 0.734 $\pm$ 0.00 & 0.928 $\pm$ 0.02 & 0.819 $\pm$ 0.01 \\
0.85 & 0.640 $\pm$ 0.02 & 0.715 $\pm$ 0.01 & 0.843 $\pm$ 0.02 & 0.774 $\pm$ 0.01 \\
\bottomrule
\end{tabular}
\end{table}
\paragraph{Effect of contrastive loss weight $\lambda_{\text{con}}$:}
Finally, we investigate the influence of the contrastive loss weighting parameter $\lambda_{\text{con}}$ on BreastMNIST clustering performance. As shown in Table~\ref{tab:ablation_lambda_breastmnist}, appropriate balancing between structural regularization and contrastive consistency is crucial for learning stable and discriminative graph representations.
\begin{table}[!htbp]
\centering
\caption{Ablation study on the effect of contrastive loss weight
$\lambda_{\text{con}}$ for the BreastMNIST dataset. Results are reported as mean
$\pm$ standard deviation over ten runs.}
\label{tab:ablation_lambda_breastmnist}
\setlength{\tabcolsep}{6pt}
\renewcommand{\arraystretch}{1.05}
\begin{tabular}{c c c c c}
\toprule
$\lambda_{\text{con}}$ & Accuracy & Precision & Recall & F1 Score \\
\midrule
0.001 & 0.731 $\pm$ 0.01 & 0.741 $\pm$ 0.00 & 0.996 $\pm$ 0.02 & 0.836 $\pm$ 0.01 \\
0.01  & 0.701 $\pm$ 0.01 & 0.729 $\pm$ 0.00 & 0.940 $\pm$ 0.02 & 0.821 $\pm$ 0.01 \\
0.5   & 0.570 $\pm$ 0.01 & 0.757 $\pm$ 0.01 & 0.607 $\pm$ 0.01 & 0.674 $\pm$ 0.01 \\
0.9   & 0.563 $\pm$ 0.02 & 0.754 $\pm$ 0.03 & 0.597 $\pm$ 0.01 & 0.666 $\pm$ 0.01 \\
2.0   & 0.570 $\pm$ 0.01 & 0.761 $\pm$ 0.01 & 0.599 $\pm$ 0.02 & 0.671 $\pm$ 0.01 \\
\bottomrule
\end{tabular}
\end{table}
\subsection{PneumoniaMNIST}

\paragraph{Effect of feature backbone.}
We compare pretrained ViT-DINO and ResNet-18 representations to evaluate the influence of the feature extraction backbone. As shown in Table~\ref{tab:ablation_backbone_pneumoniamnist}, both backbones achieve strong performance, with ResNet-18 providing slightly better overall results, suggesting effective capture of localized radiological patterns in chest X-ray images.
\begin{table}[!htbp]
\centering
\caption{Ablation study on the effect of feature extraction backbone for the
PneumoniaMNIST dataset. Results are reported as mean $\pm$ standard deviation
over ten runs.}
\label{tab:ablation_backbone_pneumoniamnist}
\setlength{\tabcolsep}{6pt}
\renewcommand{\arraystretch}{1.05}
\begin{tabular}{l c c c c}
\toprule
Backbone & Accuracy & Precision & Recall & F1 Score \\
\midrule
ViT-DINO & 0.904 $\pm$ 0.00 & 0.968 $\pm$ 0.00 & 0.856 $\pm$ 0.00 & 0.909 $\pm$ 0.00 \\
Resnet-18 & 0.910 $\pm$ 0.03 & 0.971 $\pm$ 0.03 & 0.865 $\pm$ 0.03 & 0.915 $\pm$ 0.03 \\
\bottomrule
\end{tabular}
\end{table}
\paragraph{Effect of graph sparsification threshold $\alpha$:}
We further investigate the sensitivity of PneumoniaMNIST clustering performance to the graph sparsification threshold $\alpha$. Moderate sparsification improves graph connectivity and leads to the best overall clustering performance, as shown in Table~\ref{tab:ablation_alpha_pneumoniamnist}.
\begin{table}[!htbp]
\centering
\caption{Ablation study on the effect of graph sparsification threshold $\alpha$
for the PneumoniaMNIST dataset. Results are reported as mean $\pm$ standard
deviation over ten runs.}
\label{tab:ablation_alpha_pneumoniamnist}
\setlength{\tabcolsep}{6pt}
\renewcommand{\arraystretch}{1.05}
\begin{tabular}{c c c c c}
\toprule
$\alpha$ & Accuracy & Precision & Recall & F1 Score \\
\midrule
0.80 & 0.862 $\pm$ 0.00 & 0.932 $\pm$ 0.00 & 0.812 $\pm$ 0.00 & 0.868 $\pm$ 0.00 \\
0.85 & 0.876 $\pm$ 0.00 & 0.955 $\pm$ 0.00 & 0.817 $\pm$ 0.00 & 0.881 $\pm$ 0.00 \\
0.90 & 0.910 $\pm$ 0.03 & 0.971 $\pm$ 0.03 & 0.865 $\pm$ 0.03 & 0.915 $\pm$ 0.03 \\
\bottomrule
\end{tabular}
\end{table}
\paragraph{Effect of contrastive loss weight $\lambda_{\text{con}}$:}  
We also assess how the contrastive loss weight $\lambda_{\text{con}}$ affects PneumoniaMNIST clustering performance. Table~\ref{tab:ablation_lambda_pneumoniamnist} shows that moderate contrastive regularization delivers consistently strong results across various weighting values.
\begin{table}[!htbp]
\centering
\caption{Ablation study on the effect of contrastive loss weight
$\lambda_{\text{con}}$ for the PneumoniaMNIST dataset. Results are reported as
mean $\pm$ standard deviation over ten runs.}
\label{tab:ablation_lambda_pneumoniamnist}
\setlength{\tabcolsep}{6pt}
\renewcommand{\arraystretch}{1.05}
\begin{tabular}{c c c c c}
\toprule
$\lambda_{\text{con}}$ & Accuracy & Precision & Recall & F1 Score \\
\midrule
0.005 & 0.905 $\pm$ 0.00 & 0.965 $\pm$ 0.00 & 0.861 $\pm$ 0.00 & 0.910 $\pm$ 0.00 \\
0.01  & 0.905 $\pm$ 0.00 & 0.965 $\pm$ 0.00 & 0.861 $\pm$ 0.00 & 0.910 $\pm$ 0.00 \\
5 & 0.910 $\pm$ 0.03 & 0.971 $\pm$ 0.03 & 0.865 $\pm$ 0.03 & 0.915 $\pm$ 0.03 \\
\bottomrule
\end{tabular}
\end{table}
\subsection{liver ultrasound Dataset}

\paragraph{Effect of feature backbone:}
We evaluate the influence of the feature extraction backbone by comparing pretrained ViT-DINO representations with a ResNet-18 encoder on the liver ultrasound dataset. The results highlight the importance of robust pretrained embeddings for capturing discriminative tumor patterns and improving clustering performance.
\begin{table}[!htbp]
\centering
\caption{Ablation study on the effect of feature extraction backbone for the
liver ultrasound dataset. Results are reported as mean $\pm$ standard deviation
over ten runs.}
\label{tab:ablation_backbone_liver}
\setlength{\tabcolsep}{6pt}
\renewcommand{\arraystretch}{1.05}
\begin{tabular}{l c c c c}
\toprule
Backbone & Accuracy & Precision & Recall & F1 Score \\
\midrule
ViT-DINO & 0.754 $\pm$ 0.00 & 0.835 $\pm$ 0.00 & 0.800 $\pm$ 0.00 & 0.817 $\pm$ 0.00 \\
ResNet-18 & 0.685 $\pm$ 0.00 & 0.685 $\pm$ 0.00 & 1.000 $\pm$ 0.00 & 0.813 $\pm$ 0.00 \\
\bottomrule
\end{tabular}
\end{table}
\paragraph{Effect of graph sparsification threshold $\alpha$:}
We next study the sensitivity of the liver ultrasound population graph to the
sparsification threshold $\alpha$. As shown in
Table~\ref{tab:ablation_alpha_liver}, the choice of $\alpha$ significantly
affects graph connectivity and downstream classification performance.
\begin{table}[!htbp]
\centering
\caption{Ablation study on the effect of graph sparsification threshold $\alpha$
for the liver ultrasound dataset. Results are reported as mean $\pm$
standard deviation over ten runs.}
\label{tab:ablation_alpha_liver}
\setlength{\tabcolsep}{6pt}
\renewcommand{\arraystretch}{1.05}
\begin{tabular}{c c c c c}
\toprule
$\alpha$ & Accuracy & Precision & Recall & F1 Score \\
\midrule
0.70 & 0.512 $\pm$ 0.00 & 0.651 $\pm$ 0.00 & 0.618 $\pm$ 0.00 & 0.634 $\pm$ 0.00 \\
0.83 & 0.754 $\pm$ 0.00 & 0.835 $\pm$ 0.00 & 0.800 $\pm$ 0.00 & 0.817 $\pm$ 0.00 \\
0.87 & 0.688 $\pm$ 0.00 & 0.836 $\pm$ 0.00 & 0.678 $\pm$ 0.00 & 0.749 $\pm$ 0.00 \\
0.95 & 0.666 $\pm$ 0.00 & 0.707 $\pm$ 0.00 & 0.876 $\pm$ 0.00 & 0.782 $\pm$ 0.00 \\
\bottomrule
\end{tabular}
\end{table}
\paragraph{Effect of contrastive loss weight $\lambda_{\text{con}}$:}  
Finally, we examine how the contrastive loss weight $\lambda_{\text{con}}$ affects clustering performance for the liver ultrasound dataset. The results in Table~\ref{tab:ablation_lambda_liver} show that stronger contrastive regularization improves performance, with the best results obtained at $\lambda_{\text{con}} = 8$.

\begin{table}[!htbp]
\centering
\caption{Ablation study on the effect of contrastive loss weight
$\lambda_{\text{con}}$ for the liver ultrasound dataset. Results are reported as
mean $\pm$ standard deviation over ten runs.}
\label{tab:ablation_lambda_liver}
\setlength{\tabcolsep}{6pt}
\renewcommand{\arraystretch}{1.05}
\begin{tabular}{c c c c c}
\toprule
$\lambda_{\text{con}}$ & Accuracy & Precision & Recall & F1 Score \\
\midrule
0.001 & 0.728 $\pm$ 0.06 & 0.805 $\pm$ 0.01 & 0.795 $\pm$ 0.01 & 0.800 $\pm$ 0.00 \\
0.05  & 0.726 $\pm$ 0.01 & 0.820 $\pm$ 0.01 & 0.769 $\pm$ 0.02 & 0.794 $\pm$ 0.01 \\
0.9   & 0.716 $\pm$ 0.02 & 0.822 $\pm$ 0.03 & 0.752 $\pm$ 0.07 & 0.783 $\pm$ 0.03 \\
5     & 0.730 $\pm$ 0.01 & 0.812 $\pm$ 0.03 & 0.793 $\pm$ 0.04 & 0.801 $\pm$ 0.01 \\
8     & 0.754 $\pm$ 0.00 & 0.835 $\pm$ 0.00 & 0.800 $\pm$ 0.00 & 0.817 $\pm$ 0.00 \\
\bottomrule
\end{tabular}
\end{table}

\section{Conclusion}

In this work, we proposed ARMA-C$^3$, a contrastive graph learning framework that leverages graph cut for subject-level representation learning in biomedical datasets. The proposed method models inter-subject relationships using graph structures and effectively addresses challenges such as high dimensionality, limited sample sizes, and noisy measurements by incorporating structural regularization with multi-view contrastive learning. As a result, ARMA-C$^3$ learns stable, well-separated, and robust representations.

In addition, the framework provides meaningful interpretability through histogram-based diffusion MRI features, particularly for ADNI and NIFD datasets, while also showing strong generalization across multiple imaging modalities. The overall results highlight the potential of ARMA-C$^3$ as a unified and clinically relevant framework for biomedical data analysis. In clinical pipelines where annotation is expensive (e.g., dMRI cohort studies, ultrasound triage), ARMA-C$^3$ can serve as a label-efficient pre-screening tool, reducing the burden of manual annotation while maintaining diagnostically meaningful subgroup separation.

\subsection{Limitations and Future Work}

Despite promising results, several limitations remain. First, the proposed framework is sensitive to graph construction hyperparameters, particularly the sparsification threshold $\alpha$, requiring dataset-specific tuning to balance structural preservation and noise suppression. Second, the current ADNI evaluation focuses only on binary classification settings (CN vs. MCI and CN vs. AD), and future work will investigate multi-class disease progression modeling. Third, the image-based experiments rely on pretrained ViT-DINO and ResNet feature extractors rather than fully end-to-end optimization, which may partially influence downstream performance. Finally, the current framework employs a static transductive graph construction strategy, requiring graph reconstruction when new subjects are added. Future work will explore adaptive graph learning, inductive graph neural networks, and scalable dynamic graph formulations for broader clinical applicability.

\section*{Data Availability}
Data used in the preparation of this article were obtained from the Alzheimer’s Disease Neuroimaging Initiative (ADNI) database (\url{https://adni.loni.usc.edu}) and the Neuroimaging in Frontotemporal Dementia (NIFD) study (FTLDNI) (\url{https://ida.loni.usc.edu}).

The investigators within ADNI and NIFD contributed to the design and implementation of their respective studies and/or provided data but did not participate in the analysis or writing of this report.

A complete listing of ADNI investigators can be found at:  
\url{http://adni.loni.usc.edu/wp-content/uploads/how_to_apply/ADNI_Acknowledgement_List.pdf}

Information about the NIFD (FTLDNI) study and its investigators is available at:  
\url{https://ida.loni.usc.edu/login.jsp?project=NIFD}
\section*{Conflict of Interest}
The authors declare that they have no conflict of interest.
\bibliographystyle{elsarticle-num-names}
\bibliography{references}
\end{document}